\documentclass[lettersize,journal]{IEEEtran}
\usepackage{amsmath,amsfonts}
\usepackage{algorithmic}
\usepackage{algorithm}
\usepackage{array}
\usepackage{textcomp}
\usepackage{stfloats}
\usepackage{url}
\usepackage{cite}

\usepackage{amssymb}
\usepackage{float}
\usepackage{graphicx}
\usepackage{subcaption}
\usepackage{amsmath}
\usepackage{enumerate}
\usepackage{booktabs} 
\usepackage{algorithm}
\usepackage{algorithmic}
\usepackage{setspace}
\usepackage{tabularx}

\begin{document}

\title{BRFL: A Blockchain-based Byzantine-Robust Federated Learning Model}
%
\author{Yang~Li,
	Chunhe~Xia,
	Chang~Li, 
	and~Tianbo~Wang,~\IEEEmembership{Member,~IEEE,}
	\thanks{Yang Li is with the Key Laboratory of Beijing Network Technology, Beihang University, Beijing 100191, China (e-mail: johnli@buaa.edu.cn).}
	\thanks{Chunhe Xia is with the Key Laboratory of Beijing Network Technology, Beihang University, Beijing 100191, China, and also with the Guangxi 		Collaborative Innovation Center of Multi-Source Information Integration and Intelligent Processing, Guangxi Normal University, Guilin 541004, China 		(e-mail: xch@buaa.edu.cn).}
	\thanks{Chang Li is with the School of Computer Science and Technology, Zhengzhou University of Light Industry, Zhengzhou 450000, China (e-mail: 3031169424@qq.com).}
	\thanks{Tianbo Wang is with the School of Cyber Science and Technology, Beihang University, Beijing 100191, China, and also with the Shanghai Key Laboratory 	of Computer Software Evaluating and Testing, Shanghai 201112, China (e-mail: wangtb@buaa.edu.cn). } 
	\thanks{(\textit{Corresponding author:}Tianbo Wang.)}
	\thanks{Manuscript received April 19, 2005; revised August 26, 2015.}}

\markboth{Journal of \LaTeX\ Class Files,~Vol.~14, No.~8, August~2021}%
{Shell \MakeLowercase{\textit{et al.}}: A Sample Article Using IEEEtran.cls for IEEE Journals}


\maketitle

\begin{abstract}

With the increasing importance of machine learning, the privacy and security of training data have become critical. Federated learning, which stores data in distributed nodes and shares only model parameters, has gained significant attention for addressing this concern. However, a challenge arises in federated learning due to the Byzantine Attack Problem, where malicious local models can compromise the global model's performance during aggregation. This article proposes the \underline{B}lockchain-based Byzantine-\underline{R}obust \underline{F}ederated \underline{L}earning (BRLF) model  that combines federated learning with blockchain technology. This integration enables traceability of malicious models and provides incentives for locally trained clients. Our approach involves selecting the aggregation node based on Pearson's correlation coefficient, and we perform spectral clustering and calculate the average gradient within each cluster, validating its accuracy using local dataset of the aggregation nodes. Experimental results on public datasets demonstrate the superior byzantine robustness of our secure aggregation algorithm compared to other baseline byzantine robust aggregation methods, and proved our proposed model effectiveness in addressing the resource consumption problem.

\end{abstract}

\begin{IEEEkeywords}
Federated Learning, Blockchain, Pearson Correlation Coefficient, Spectral Clustering
\end{IEEEkeywords}

\section{Introduction}
\IEEEPARstart{F}{ederated} learning \cite{McMahan2016CommunicationEfficientLO} (FL) has revolutionized the traditional approach to machine learning by storing data locally on multiple nodes rather than on centralized servers. These nodes train the same model and then upload it to the server, allowing devices like smartphones, laptops, and in-car computers to participate in model training.

However, federated learning has both positive and negative aspects \cite{Li2019FederatedLC}. On the positive side, it enhances the privacy and security of training data by keeping it locally stored and not shared with the server. It finds applications in various domains such as transportation, healthcare, and finance. On the negative side, since training data is not stored on the server, the server needs to aggregate parameters from multiple local models, which can be vulnerable to attacks from malicious nodes. Research in robust global model aggregation focuses on situations where some local models are under attack, while maintaining high performance for the global model. The key aspect of this research is to identify the attacked models and exclude them during the global model aggregation process, while retaining the benign models for aggregation. Therefore, secure global model aggregation algorithms play a vital role in federated learning. Specifically, our research aims to combat malicious local models and trace their activities. According to the survey \cite{Qu2022BlockchainenabledFL}, the detection of malicious models can be achieved through model aggregation algorithms, and when combined with blockchain technology, it enables traceability of malicious gradients. Additionally, it provides rewards to nodes participating in federated learning, mitigating potential attacks such as middleman attacks and single point failures, and increasing the enthusiasm of nodes to participate in federated learning training.

In blockchain-based federated learning, there are two inevitable problems in model aggregation: \textit{byzantine-attack} and \textit{resource-cost}. These problems arise because it is not guaranteed that every local model trains in the intended direction. Local models are sent to the server for aggregation, but the server does not participate in their training. As a result, the server cannot verify the validity of the local models it receives, as it cannot access the local training data.  \textit{Byzantine-attack} refer to the modification of training data or model parameters by local training nodes to attack the global model. The goal of such attacks is to reduce the accuracy of the global model or manipulate predictions for specific data. \textit{Resource-cost} refer to the significant computational resources consumed by existing secure aggregation methods that rely on local model accuracy verification. Additionally, solutions combining blockchain technology face challenges related to block storage.

Researchers have developed aggregation algorithms mainly based on $l_p$ distance \cite{Chen2017DistributedSM, Cao2019UnderstandingDP}, cosine angle \cite{Khazbak2020MLGuardMP, Fung2020TheLO}, and accuracy validation \cite{Wang2020ModelPD, Tan2020TowardSS, Chen2021ZeroKC} to detect malicious local models for secure model aggregation in federated learning. The $l_p$ distance-based method calculates the distance between local model gradients and assumes that the distance between malicious model gradients and benign model gradients is larger than the distance between benign models, thereby detecting malicious models. The cosine angle-based method considers the angle between local models and previous global models, identifying them as malicious if the cosine angle exceeds a certain threshold. The accuracy validation-based method verifies the accuracy of local models using a prepared test dataset and identifies them as malicious if their accuracy falls below a certain threshold. 

Unfortunately, these three primary aggregation algorithms do not effectively address the issue of detecting malicious models. The $l_p$ distance-based method can be vulnerable to attackers who add little biases to the parameters of normal gradients, thereby attacking the global model. In the cosine angle-based method, attackers can manipulate the angles between malicious model updates and benign model updates to be smaller than the discrimination threshold of the defense method used. As for the accuracy validation-based method, it requires an additional test dataset and significant computational resources to validate the accuracy of local models. All three methods assume that the proportion of malicious models is less than 50\%.

To address these issues, we propose the Blockchain-based Federated Learning model (BRFL), which contains two main components: the Proof of Pearson Correlation Coefficient consensus algorithm (PPCC) and the Precision-based Spectral Aggregation  (PSA)  algorithm. PPCC selects the aggregation node for the next round based on the Pearson correlation coefficient between the local model and the global model from previous rounds. It also considers the lack of test datasets in FL by using the aggregation node's local dataset for local model accuracy validation. PSA clusters local models with high correlation and verifies their accuracy by calculating the average value, thereby detecting malicious models and addressing the resource cost problem. Experimental results demonstrate that our model exhibits high robustness and effectively reduces resource consumption.

Our contributions can be summarized as follows.
\begin{enumerate}[1.]
	
	\item We propose the BRFL model to address the issue of \textit{byzantine-attack} during the model aggregation process. We propose a new block structure by adding a transaction type identity in the block head and utilizing IPFS for storage of local models. This allows nodes to save the chosen transaction type of the block.
	
	\item In BRFL, we introduce a consensus algorithm suitable for blockchain-based federated learning. The algorithm selects the node with the highest Pearson correlation coefficient between the previous round's local model and the global model as the aggregation node. This approach enhances the security of federated learning and resolves the problem of lacking test data in FL.
	
	\item We propose a Precision-based Spectral Aggregation (PSA) algorithm , which leverages spectral clustering of local models based on their Pearson correlation coefficients. This algorithm verifies the accuracy of the average gradient within each cluster using the aggregation nodes' local datasets, selects the average gradient from the cluster with higher accuracy as the global model. This approach ensures that the global model maintains high accuracy even in the presence of a high percentage of malicious models.

	\item Experiments conducted on two commonly used datasets demonstrate that our aggregation algorithm surpasses both classic and state-of-the-art baseline methods in terms of accuracy. Additionally, our block structure effectively reduces the storage pressure on federated learning nodes. The incentive mechanism successfully rewards well-intentioned nodes, and we validate the effectiveness of the consensus algorithm.
\end{enumerate}

The rest of this article is organized as follows. We first give a comprehensive review of related works in Section \ref{RELATED WORK}. Next, we demonstrate the background of this article in Section \ref{BACKGROUND},
followed by the presentation of  detailed design of BRFL in Section \ref{Blockchain-based Federated Learning Model}. After that, we conduct a series of experiments on two public datasets to evaluate BRFL in Section \ref{PERFORMANCE EVALUATION}. Finally,
Section  \ref{Conclusion} concludes this work and discusses future directions.

\section{RELATED WORK}
\label{RELATED WORK}
This section provides a comprehensive review of the relevant literature. We discuss two approaches to enhancing the security of federated learning: combining blockchain and Byzantine robust aggregation methods.

Federated learning and blockchain share a distributed architecture. Currently, there are solutions that improve the security of federated learning by using improved consensus algorithms, distributed ledgers, and encryption algorithms in blockchain. Bao \MakeLowercase{\textit{et al.}}  \cite{Bao2019FLChainAB}  proposed FLChain, where blocks in FLChain store participants' personal information, data resource descriptions, historical federal training activities, and credibility, allowing historical information of federated learning to be queried in the blockchain.  The dual committee introduced by \cite{Lin2022ABD}, ensures the selection of good nodes and other nodes to build the committee, enabling more nodes to participate and preventing malicious nodes from attacking the training process. Islam \MakeLowercase{\textit{et al.}}  \cite{Islam2022ABP} presents a blockchain-based data acquisition scheme during the pandemic, where federated learning is used to assemble privacy-sensitive data as a trained model instead of raw data. They also design a secure training scheme to mitigate cyber threats. Bift, proposed by  \cite{He2022BiftAB} , is a fully decentralized ML system combined with FL and blockchain. It provides a privacy-preserving ML process for connected and autonomous vehicles, allowing them to train their own machine learning models locally using their driving data and then upload the models to obtain a better global model. Biscotti, proposed by \cite{Shayan2021BiscottiAB} , is a fully decentralized peer-to-peer (P2P) approach to multi-party machine learning. It utilizes blockchain and cryptographic primitives to coordinate a privacy-preserving ML process between peering clients, ensuring scalability, fault tolerance, and defense against known attacks. Liu \MakeLowercase{\textit{et al.}} \cite{Liu2021BlockchainAF} proposed a cooperative intrusion detection mechanism that unloads the training model to distributed edge equipment, such as connecting vehicles and roadside units. This method reduces the resource utilization of the central server while ensuring security and privacy. To ensure the security of the aggregation model, blockchain is used to store and share training models. 
Chen \MakeLowercase{\textit{et al.}} \cite{Chen2021RobustBF} proposed a blockchain-based decentralized FL framework,  introduce a novel decentralized verification mechanism,  design a specialized proof-of-equity consensus mechanism where honest devices can be rewarded with equity more frequently, thus protecting legitimate local model updates by increasing their chances of dominating blocks attached to the blockchain.

Several Byzantine robust methods have been proposed. Yuan \MakeLowercase{\textit{et al.}} \cite{Yuan2022ABD} provides a problem formulation of vertical federated learning (FL) in the presence of \textit{byzantine-attack} and proposes a Byzantine-resilient dual subgradient method. Chen \MakeLowercase{\textit{et al.}} \cite{Chen2021BDFLAB} proposes BDFL, a novel Byzantine fault-tolerant decentralized FL method, to provide privacy protection for self-driving cars. By extending the HydRand protocol, a point-to-point FL with Byzantine fault tolerance is built. To protect their models, each autonomous vehicle (AV) uses a publicly verifiable secret sharing scheme that allows anyone to verify the correctness of the encrypted sharing. Bao \MakeLowercase{\textit{et al.}} \cite{Bao2022BOBABF} proposes BOBA, an efficient two-stage method that aims to solve label skewness and ensure robustness, along with efficient optimization algorithms. Chen \MakeLowercase{\textit{et al.}} \cite{Chen2022ByzantineresilientFL} introduces a new Gradient Average framework, which utilizes the average value of multiple gradients to detect \textit{byzantine-attack}. The AG framework can be combined with most existing defenses, such as Krum \cite{Blanchard2017MachineLW}, to reduce variance and improve effectiveness.  Jin \MakeLowercase{\textit{et al.}} \cite{Jin2022ByzantineRobustAE} proposes a new aggregation rule for FL called Federal Mutual Assistance Information (FedMI), which utilizes mutual information between customers to build resilience for Byzantine workers and accelerate integration. Li \MakeLowercase{\textit{et al.}}  \cite{Li2022CommunicationEfficientAB} presents a new method called PCS-DP-SIGNSGD, which utilizes a simple and efficient 1-bit compressed sensing reconstruction algorithm for aggregated gradients. The accuracy loss caused by the recovery error is almost negligible.

\section{BACKGROUND}
\label{BACKGROUND}

\subsection{Blockchain}

Blockchain \cite{Nakamoto2008BitcoinAP}, proposed by Nakamoto in 2008, aims to establish a decentralized and untrusted digital cash system. It is based on a peer-to-peer network and incorporates encryption algorithms, consensus algorithms, and other technologies.

In traditional cash systems, an institution endorses the currency, thereby assigning value to the coin. People recognize the value of the currency by acknowledging the credibility of the endorsing organization. However, in the first blockchain application, Bitcoin, Nakamoto introduces a method for everyone to maintain records of digital cash transactions between any individuals, eliminating the need for a single organization to endorse the trust of the transaction. The more people involved in ledger-keeping, the more secure and reliable the transaction history becomes.

A blockchain consists of blocks that are connected through a bidirectional linked list. Each block is divided into a block head and a block body. The block head contains the block generation timestamp, the hash value of the previous block, the hash value of the current block, the Merkle root (which ensures transaction consistency within the block), the block height, and other information. The block body contains transactions that occurred during a specific time period. Once recorded in a block, the transactions cannot be modified.

Consensus algorithms \cite{Xiong2022ResearchOP} play a crucial role in blockchain by organizing network nodes to reach a consensus on transactions and ensuring the correctness and continuity of the blockchain. In Bitcoin, the Proof of Work (PoW) consensus algorithm was introduced, where miners compete to solve a complex mathematical puzzle to generate the next block. The first miner to solve the puzzle successfully adds the block to the chain and receives rewards for their efforts.

Blockchain can be utilized to address privacy protection issues in federated learning \cite{Farooq2022BlockchainFL, Aich2021ProtectingPH, Jiao2023ABF }. By leveraging the tamper-proof, transparent, and decentralized nature of blockchain, the security of the federated learning system can be enhanced by storing important data on the blockchain.

\subsection{Pearson Correlation Coefficient}

Pearson correlation coefficient (PCC) \cite{PearsonIIICT} aims to describe the linear correlation between two variables, assume there has a variable $A = \{a_1, a_2, …, a_n\}$, $B = \{b_1, b_2, \ldots , b_n\}$, the PCC score $\rho$ can be calculate as :

\begin{equation}
	\begin{aligned}
		& \rho(A,B) \\
		& = \frac {E[(A-\mu_A)(B-\mu_B)]} {\sigma_A\sigma_B} \\
		&=\frac{E(A,B)-E(A)E(B)}{\sqrt{(E(A^2 )-E^2 (A) )} \sqrt{(E(B^2 )-E^2 (B) )}} \\
		&=\frac{n\sum\limits_{i=1}^{n} a_i b_i - (\sum\limits_{i=1}^{n} a_i)(\sum\limits_{i=1}^{n} b_i)} {\sqrt{n\sum\limits_{i=1}^{n} a_i^2 - (\sum\limits_{i=1}^{n} a_i)^2} \sqrt{n\sum\limits_{i=1}^{n} b_i^2 - (\sum\limits_{i=1}^{n} b_i)^2}} \\
	\end{aligned}
\end{equation}

Hence, $\rho(A,B) \in [-1,1]$, $\rho(A,B) \leftarrow 0$ means variable $A$ and $B$ has no linear correlations, $\rho(A,B) \leftarrow 1$ means $A$ and $B$ has positive linear correlations, when one of them increase, the other will also increase; $\rho(A,B) \leftarrow -1$ means $A$ and $B$ has negative linear correlations, when one of them increase, the other will decrease. PCC has features, such as:
\begin{enumerate}[1)]
	\item Symmetry. $\rho$(A,B) = $\rho$(B,A), the PCC between the two specific variables is the same.
	\item Displacement invariant. Changes in the expected values of variable $A$ and variable $B$ do not affect the PCC. 
	\item Scale invariance. Exponential change in variable $A$ and variable $B$ does not affect the PCC.
\end{enumerate}
PCC can be used to describe two vectors relation. For the models  in federated learning, the PCC can be used to describe the strength and direction of the linear relationship between every two models.

\subsection{Spectral Clustering}

The fundamental concept of spectral clustering \cite{Luxburg2007ATO} is to convert samples and their similarity matrix into a graph model. By cutting the edges of the graph model, it aims to minimize the edge weights between different subgraphs and maximize the edge weights within each subgraph. Spectral clustering is employed to classify samples and tackle the clustering problem of sparse data, which is a limitation of the k-means clustering algorithm.

Assuming we have a dataset that can be classified into $k$ classes, $X = \{x_1,x_2,...,x_n \} \in \mathbb{R}^{d \times n}$, the algorithm proceeds as follows:
\begin{enumerate}
	\item Construct the similarity matrix $S \in \mathbb{R}^{n \times n}$ and obtain the corresponding degree matrix $D \in \mathbb{R}^{n \times n}$, adjacency matrix $W \in \mathbb{R}^{n \times n}$, and Laplacian matrix $L = D - W$.
	\item Normalize $L$ by $L^*=D^{-1/2}LD^{-1/2}$ and compute the $m$ smallest eigenvalues of $L^*$ and their corresponding eigenvectors $f$.
	\item The eigenvector matrix $F$, composed of the corresponding eigenvectors $f$, Normalize $F$ row-wise to obtain the final $n \times m$-dimensional feature matrix $ F \in \mathbb{R}^{n \times m}$.
	\item Cluster the row vectors of $F$, where each row vector represents an $m$-dimensional sample. With a total of $n$ samples, set the number of clusters as $k$, and obtain the partitioning of clusters $C = \{c_1,c_2,...,c_k\}$.
\end{enumerate}

Characteristics of spectral clustering:
\begin{enumerate}
	
	\item Spectral clustering only requires the similarity matrix of the data, making it effective for clustering sparse data.
	
	\item Due to its utilization of dimensionality reduction techniques, spectral clustering exhibits better complexity than traditional clustering algorithms when dealing with high-dimensional data.
\end{enumerate}

\section{Blockchain-based Federated Learning Model}
\label{Blockchain-based Federated Learning Model}

In this section, we present a novel approach that combines blockchain technology with federated learning to enhance security and robustness. We propose BRLF, which incorporates a customized consensus algorithm and a robust aggregation mechanism. Our approach leverages blockchain technology, Pearson correlation coefficient (PCC), and spectral clustering techniques to effectively filter out malicious gradients during the global model aggregation process. This approach offers enhanced security and robustness for federated learning while reducing computation power costs. Furthermore, we introduce the utilization of IPFS and a new block structure to minimize storage resource costs.

\subsection{Model overview}

\begin{figure*}[htbp]
	\centering
	\includegraphics[width=0.6\textwidth]{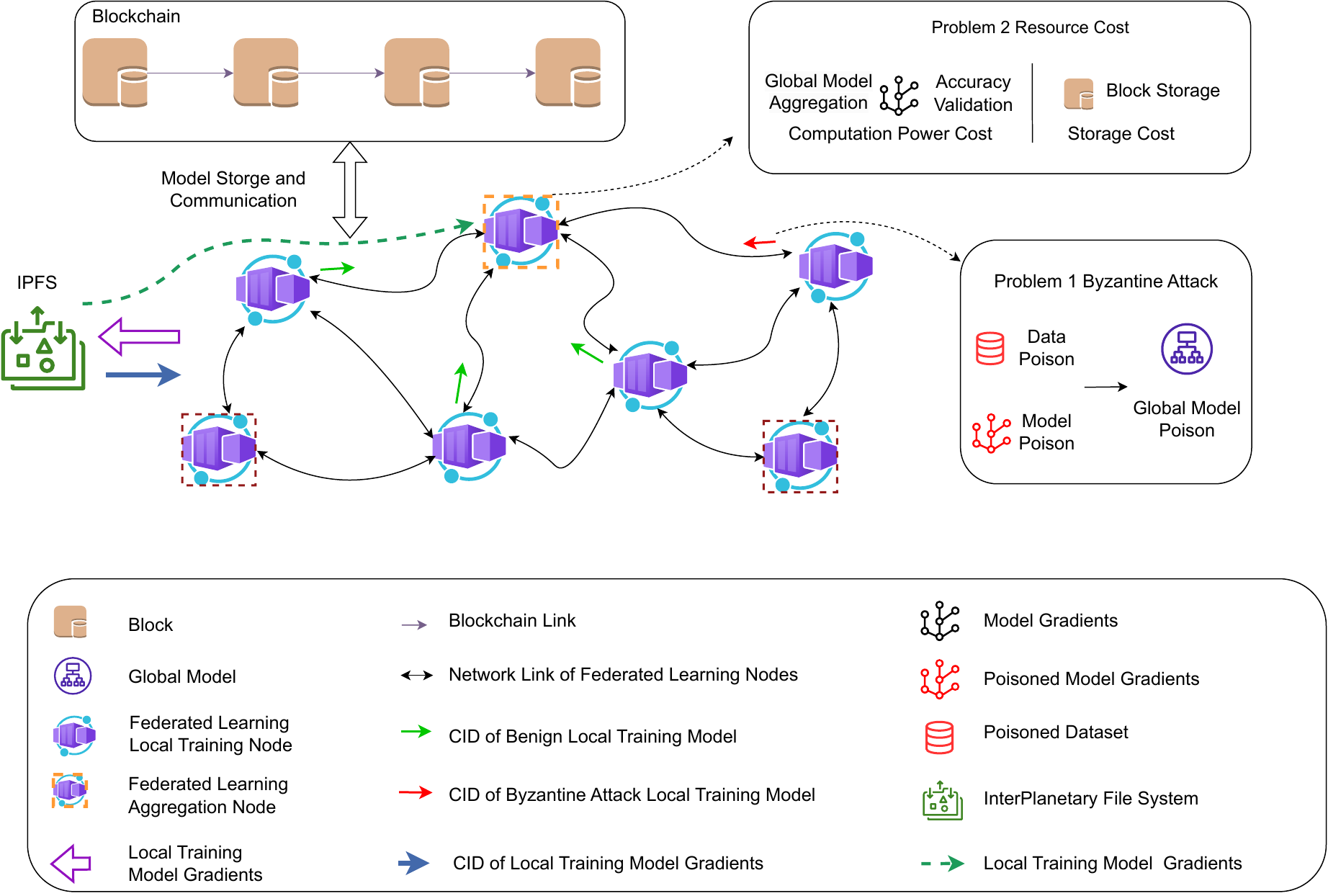}
	\caption{BRFL Model Structure }
\end{figure*} 

In figure 1 shows the architecture of BRFL. In order to improve the security of global model aggregation, BRFL design goal are as followings:

\begin{itemize}
	
	\item Aggregating local update model gradients makes the global model resilient against \textit{byzantine-attack}, ensuring that the aggregated global model performs well even in the presence of such attacks.
	
	\item By incorporating  consensus algorithms in blockchain, the selection of aggregation nodes is guaranteed to maintain the quality and efficiency of global model aggregation.
	
	\item Blockchain technology is utilized to achieve secure transmission and traceability of model updates, ensuring that the transmission process remains untampered and providing the ability to trace malicious gradients.
\end{itemize}

As depicted in Figure 1, the local model and global model are stored in the blockchain. Local training nodes retrieve the global model from the blockchain and utilize their own datasets to train the local model. After local training, the updated local model gradient is uploaded to IPFS, and the Content Identifier (CID) returned by IPFS is then uploaded to the blockchain. 

Let's assume there are $\nu$ clusters ($\nu \in \mathbb{N}^+$), and the average of the local model in the $\lambda$-th cluster, denoted as $CL_{a;\lambda}^{lm}$. Since local models within the same cluster exhibit similar linear correlations, $CL_{a;\lambda}^{lm}$ can effectively represent the model distribution in that cluster. $CL_{a;\lambda}^{lm}$ is verified by all the aggregation nodes, and the sum of the average accuracy from each node is considered the final accuracy, as test results from a single aggregation node may be biased. The aggregation nodes are selected by comparing the PCCs of the last round of local and global models
. It is important to note that the aggregation nodes from the previous round will not be selected as aggregation nodes for the next round. This is because aggregation nodes are not involved in training the local model but rather focus on verifying the local models and packaging blocks. The immutability of the blockchain ensures that the global model, local model, accuracy records, and Pearson correlation coefficient records cannot be tampered with, thus enhancing the security of federated learning. 

$$CL_{a;\lambda}^{lm} = \frac{\sum\limits_{\mu = 1}^{\omega} lm_{\mu}}{\omega} $$

Hence, let $\omega \in \mathbb{N}^+$ and $\omega < \beta$, where $\beta$ is a dynamic number that indicates the number of nodes in each cluster.

In BRFL, 
  we have added a signal in the block head to indicate the type of content. This allows nodes to store only the blocks they need, thereby conserving storage resources. The transmission of local models and global models occurs through the blockchain's peer-to-peer (P2P) network. The federated learning nodes are also part of the blockchain network. Local training nodes and aggregation nodes are rewarded with tokens for their contributions to model training, aggregation and blockchain packaging.

\subsection{Initialization}

In BRFL we denote all nodes' set is $N = \{n_{1}, n_{2}, \ldots, n_{\alpha}\}, \alpha \in \mathbb{N}^+$ is the number of all nodes. $N$ contains aggregation nodes set, local training nodes set and blockchain nodes set. There has three type of nodes in BRFL:
\begin{itemize}
	
	\item Aggregation node ($An$): The set of aggregation nodes is denoted as $AN = \{An_{1},An_{2}, \ldots , An_{\beta}\}$, where $\beta \in \mathbb{N}^+$. These nodes aggregate local models to form the global model and package blocks.
	
	\item Local training node ($Lt$): The set of local training nodes is denoted as $LT = \{Lt_{1}, Lt_{2},\ldots, Lt_{\gamma}\}$, where $\gamma \in \mathbb{N}^+$. The local training nodes are responsible for training the local models and sending them to the aggregation nodes ($An$) for the aggregation of the global model. Both the local training nodes ($Lt$) and the aggregation nodes ($An$) have their own datasets.
	
	\item Blockchain network node ($Bn$): The blockchain network nodes ($Bn$) do not participate in the federated learning process. They are denoted as $BN = {Bn_1, Bn_2, \ldots, Bn_{\varkappa}}$, where $\varkappa \in \mathbb{N}^+$. The presence of more nodes in the blockchain network ensures that the ledger is consensus by a larger number of nodes, thereby enhancing the security of the blocks in the blockchain.
\end{itemize}

There are two types of models in BRFL:

\begin{itemize}
	
	\item Local model ($lm$): The local model set is denoted as $LM = \{lm_{1}^{r}, lm_{2}^{r}, \ldots, lm_{\gamma}^{r}\}$, where $r \in \mathbb{N}^+$ represents the global training round, and $\gamma$ represents the node id. $lm$ refers to the local training model ($Lt$) and in each communication round. After local training, the updated $lm$ is published on IPFS.
	
	\item Global model ($gm$): The global model set is denoted as $GM = \{gm_{1}^{\theta}, gm_{2}^{\theta}, \ldots, gm_{\chi}^{\theta} \}$, where $\chi \in \mathbb{N}^+$ represents the global round number and $\theta$ is the aggregation node ID. The global model is aggregated by the aggregation nodes ($An$).  The global model represents a comprehensive representation of the collective information within the federated learning network, combining the data characteristics and insights contributed by each participant.
\end{itemize}

In the beginning of federated learning in BRFL, several steps are followed. First, the maximum training round ($MTR$) is set, along with the model structure. The global model parameters are randomly initialized. Next, a certain number ($\beta$) of $LT$  are randomly selected to become  $AN$. The global model is then packaged and stored in the blockchain.

\subsection{Local Model Training and Communication}

Blockchain possesses characteristics such as transparency, immutability, and decentralized maintenance. These features ensure the security and transparency of data on the blockchain, making it accessible to all nodes. However, these advantages also introduce certain challenges, including storage space and communication efficiency issues.

In a blockchain, blocks need to be stored on nodes. The more nodes that store blocks in the blockchain network, the higher the security level and the lower the likelihood of tampering.  However, storing blocks requires local storage space from the nodes. In the case of federated learning, where there are numerous training rounds, multiple participating nodes, and long training cycles, the types of participating nodes vary greatly. These nodes can include mobile phones, cars, computers, and more, each with different storage capacities.

To address these challenges, we storing the local models in the InterPlanetary File System (IPFS), while storing the pointers returned by IPFS within the blockchain. In federated learning, with multiple participating local training nodes, storing all the local models on the blockchain would result in numerous blocks generated in each training round, which hampers communication efficiency. IPFS will return a hash sequence, Content Identifier (CID), for each certain file upload, aggregation nodes can retrieve the local models from IPFS based on the CID, perform model aggregation, and store the aggregated global model within the blockchain. By storing the global model only once per round within the blockchain, local training nodes can directly access it from the blockchain instead of IPFS, thereby improving their work efficiency. Assume current round is $\varphi$ th round,  $Lt_{\varpi}, \varpi \in [1, \gamma]$ get  $gm_{\varphi - 1}^{\theta}$ from blockchain, set  $gm_{\varphi - 1}^{\theta}$ as its local model $lm_{\varpi}^{r} \leftarrow gm_{\varphi - 1}^{\theta}$, doing local training using its own dataset and computation power, get updated local model $lm_{u;\varpi}^{r}$. 
$Lt_{\varpi}$ upload their $LM$ to IPFS, get CID returns from IPFS   for the uploaded $lm_{u;\varpi}^{r}$, denoted as $CID_{lm_{u;\varpi}^{r}}$, which can be used to retrieve $lm_{u;\varpi}^{r}$. $LT$ publish $CID_{lm_{u;\varpi}^{r}}$ as a transaction to blockchain network.
\subsection{Proof of Pearson Correlation Coefficient Consensus Algorithm}

In BRFL, we propose a consensus algorithm called Proof of Pearson Correlation Coefficient (PPCC) to select the $LT$ that will become  $AN$ at the beginning of each communication round. In federated learning, one challenge is that training data is stored in training nodes and not shared with others to protect privacy. While this enhances data security, it also poses a problem for the server to verify the validity of local models, leading to potential issues such as \textit{byzantine-attack} including data poisoning and model poisoning attacks.

To address this challenge, we select a committee of  $AN$ from the $LT$ in each training round. These selected aggregation nodes use their local datasets to verify the validity of the local models from the participating training nodes. This approach helps to overcome the lack of a test dataset and improves the overall security and integrity of the federated learning process.

Assume current round is $\epsilon(\epsilon > 1)$ th round,  the algorithm details shows in Algorithm \ref{PPCC consensus algorithm}, the steps of PPCC can be summarized as  followings:

\begin{enumerate}
	\item  Concatenate last round's $LM$ and $gm$ different layer to one vector, calculate the PCC between $LM$ and $gm$;

	\begin{equation}
		\Theta( lm_{\eta}^{\epsilon - 1}) = \bigoplus_{layer_{\varrho} \in lm_{\eta}^{\epsilon - 1}} layer_{\varrho}
	\end{equation}
	
	\begin{equation}
		\Theta(gm_{\epsilon-1}^{\theta}) = \bigoplus_{layer_{\varrho} \in gm_{\epsilon-1}^{\theta}} layer_{\varrho}
	\end{equation}

	\begin{equation}
		\sigma(	\Theta (lm_{\eta}^{\epsilon - 1})) = \sqrt{E((	\Theta (lm_{\eta}^{\epsilon - 1}))^2 )-E^2 (	\Theta (lm_{\eta}^{\epsilon - 1})) }
	\end{equation}
	
	\begin{equation}
		\sigma(\Theta(gm_{\epsilon-1}^{\theta})) = \sqrt{E((\Theta(gm_{\epsilon-1}^{\theta}))^2 )-E^2 (\Theta(gm_{\epsilon-1}^{\theta})) }
	\end{equation}
	
	\begin{equation}
		\rho(lm_{\eta}^{\epsilon - 1}, gm_{\epsilon-1}^{\theta})  
		= \frac{cov(lm_{\eta}^{\epsilon - 1},gm_{\epsilon-1}^{\theta})}{\sigma(	\Theta (lm_{\eta}^{\epsilon - 1})) \sigma(\Theta(gm_{\epsilon-1}^{\theta}))}
	\end{equation}

	Hence, $lm_{\eta}^{\epsilon - 1}, \eta \in [1, \gamma]$ is the $\eta$ th local model in $\epsilon - 1$ round,  $gm_{\epsilon-1}^{\theta} $is the global model aggregate by $Lt_{\theta},\theta \in [1, \gamma]$ in $\epsilon-1$ round, $layer_{\varrho}$ is the $\varrho$ th layer in the model gradients. $\Theta( lm_{\eta}^{\epsilon - 1})$ is concatenate different layers in $lm_{\eta}^{\epsilon - 1}$ to one vector , so as $\Theta(gm_{\epsilon-1}^{\theta})$.

	\item  Sort the $\rho$ by descending order, get first $\beta$ number of $lm$, $Lt$ who upload these $lm$ become $AN$ this round.
	
	\begin{equation}
		AN \leftarrow Sort(\rho(lm_{\eta}^{\epsilon - 1}, gm_{\epsilon-1}^{\theta}) \mid \eta \in [ 1, \gamma])[:\beta]
	\end{equation}
\end{enumerate}

The selected aggregation nodes ($AN$) verify the accuracy of the local models ($lm$) and package the transactions to the blockchain. 

The block will be packaged by one $An$, but all the $AN$ in this round will be considered the block miner. Therefore, a block may have multiple miners, and all the miners will receive tokens as rewards.
\begin{algorithm}
	\renewcommand{\algorithmicrequire}{\textbf{Input:}}
	\renewcommand{\algorithmicensure}{\textbf{Output:}}
	\caption{PPCC consensus algorithm}
	\label{PPCC consensus algorithm}
	\begin{algorithmic}[1]
		\REQUIRE gm : last round global model; num\_AN: AN number;
		\STATE Initialization: PL : Pearson Correlation Coefficient list ;
		
		\IF{gm == None}{
			\RETURN random select AN from LT;
		}
		\ENDIF
		
		\FOR{each layer in gm}
		\STATE $Vector_{gm}$ $\gets$ $Vector_{gm}$.concatenate(layer);
		
		\ENDFOR
		\FOR{i = 0; i \textless LT.length; i++}
		\FOR{each layer in LT[i]}
		\STATE $Vector_{LT[i]}$ $\gets$ $Vector_{LT[i]}$.concatenate(layer);
		\ENDFOR
		
		\STATE $\rho =  E[(Vector_{gm}-\mu_{Vector_{gm}})(Vector_{LT[i]}-\mu_{Vector_{LT[i]}})] /  \sigma_{Vector_{gm}}\sigma_{Vector_{LT[i]}}$;
		
		\STATE PL[i] = $\rho$;
		\ENDFOR
		
		\STATE sorted\_dict $\gets$ sorted(PL) in descend order;
		\STATE AN $\gets$ sorted\_dict[:num\_AN]
		\ENSURE  AN ; PL
	\end{algorithmic}  
\end{algorithm}

\subsection{Precision-based Spectral Aggregation  Algorithm }

BRFL leverages IPFS to store the $LM$ and stores the CID returned by IPFS in the blockchain. This approach addresses the challenge of storing a large number of local models when there are numerous local training nodes ($LT$) in the network. The aggregation nodes ($AN$) can retrieve the corresponding $LM$ using the CID stored in the blockchain.

Existing security aggregation algorithms often identify \textit{byzantine-attack} based on factors such as the distance between local models, the accuracy of local models, and the cosine angle between local and global models. However, these methods have limitations including low fault tolerance, high resource consumption, and insensitivity to a small number of features. In this work, we propose a secure aggregation algorithm called the Precision-based Spectral Aggregation (PSA) algorithm. It combines the Pearson Correlation Coefficient (PCC), spectral clustering, and accuracy verification to detect attacked $lm$.

First, we concatenate local model and global model 's different layer to one vector, calculate the PCC between local models, as shown in Equation \ref{Theta3}, \ref{Theta4}, \ref{Theta5}, \ref{Theta6}, \ref{Theta7}. Second, we perform spectral clustering on the  local models based on PCC. This allows us to calculate the average gradient for each cluster after clustering. The aggregation nodes selected by the PPCC consensus algorithm then verifies the accuracy rate of the average gradient using its local database. The average gradient which has the max average accuracy validated by aggregation nodes will be  the global model. This approach enhances the fault tolerance, reduces resource consumption, and improves sensitivity to small sample compared to existing methods. The algorithm details shows in Algorithm \ref{PSA aggregation algorithm}.

\begin{equation}
	\label{Theta3}
	\Theta( lm_{\eta}^{\epsilon})= \bigoplus_{layer_{\varrho} \in lm_{\eta}^{\epsilon}} layer_{\varrho}
\end{equation}

\begin{equation}
	\label{Theta4}
	\Theta(lm_{\o}^{\epsilon})=\bigoplus_{layer_{\varrho} \in lm_{\o}^{\epsilon}} layer_{\varrho}
\end{equation}

\begin{equation}
	\label{Theta5}
	\sigma(	\Theta (lm_{\eta}^{\epsilon})) = \sqrt{E((	\Theta (lm_{\eta}^{\epsilon}))^2 )-E^2 (	\Theta (lm_{\eta}^{\epsilon})) }
\end{equation}

\begin{equation}
	\label{Theta6}
	\sigma(\Theta(lm_{\o}^{\epsilon})) = \sqrt{E((\Theta(lm_{\o}^{\epsilon}))^2 )-E^2 (\Theta(lm_{\o}^{\epsilon})) }
\end{equation}

\begin{equation}
	\label{Theta7}
	\begin{aligned}
		\rho(lm_{\eta}^{\epsilon}, lm_{\o}^{\epsilon}) 
		= \frac{cov(lm_{\eta}^{\epsilon},lm_{\o}^{\epsilon})}{\sigma(	\Theta (lm_{\eta}^{\epsilon})) \sigma(\Theta(lm_{\o}^{\epsilon}))} \\ 
	\end{aligned}		
\end{equation}

\begin{equation}
	\label{matrix}
	\mathbf{\rho^{\epsilon}_{L,L}} = 
	\begin{bmatrix}
		1 & pcc_{lm_1, lm_2} & \dots & pcc_{lm_1, lm_\gamma} \\
		pcc_{lm_2, lm_1} & 1 & \dots & pcc_{lm_2, lm_\gamma} \\
		\vdots & \vdots & \ddots & \vdots \\
		pcc_{lm_\gamma, lm_1} & pcc_{lm_\gamma, lm_2} & \dots & 1\\
	\end{bmatrix}
\end{equation}

Hence, $\o \in [1, \gamma] \cap \o \neq \eta$, $\mathbf{\rho^{\epsilon}_{L,L}} \in \mathbb{R}^{(\gamma, \gamma)}$ in Equation \ref{matrix} represents the matrix of Pearson Correlation Coefficient (PCC) relationships between local models in round $\epsilon$.

The strength of the linear correlation between two local models ($lm$) can be determined by their PCC value. By clustering the local models based on PCC, we can group together models that exhibit strong linear correlation. Existing security aggregation algorithms have proposed schemes to verify the accuracy of local models to detect malicious attacks. However, individually verifying the accuracy of each local model can consume a significant amount of computational resources of the aggregation nodes ($AN$). Moreover, the computational capacity of the $AN$ limits the application of such schemes, and if the local training process generates a large number of models, the caliPSAtion time will also increase.

Therefore, after clustering the local models, we only verify the accuracy of the cluster average model gradient denoted as $ACC_{a;\lambda}^{CL}$, where $\lambda \in [1,\nu]$, using the $AN$. Here, $CL_{a;\lambda}^{lm}$ represents the linear trend of the models in cluster $\lambda$. By evaluating $CL_{a;\lambda}^{lm}$, we can verify the nodes in that cluster without consuming excessive computational resources. The $CL_{a;\lambda}^{lm}$ with higher accuracy will be selected as the global model in the current round, and the node with $lm$ in the cluster with higher accuracy will receive stake rewards ($Lt$).

Global model aggregation mechanism based on conventional federated learning methods Fedavg:

$$
gm_{\epsilon}^{\theta} = \frac{\sum\limits_{\lambda = 1}^{\nu}  CL_{a;\lambda}^{lm} }  {\nu}
$$

\begin{algorithm}
	\renewcommand{\algorithmicrequire}{\textbf{Input:}}
	\renewcommand{\algorithmicensure}{\textbf{Output:}}
	\caption{PSA Algorithm.}
	\label{PSA aggregation algorithm}
	\begin{algorithmic}[1]
		\REQUIRE LMS : local model set; Num\_{cluster} : cluster number; LT : local training nodes set; AN : aggregation nodes set;
		\STATE Initialization: $PM$: Pearson Correlation Coefficient matrix ; $AL$ : accuracy validation list;
		
		\FOR{i = 0; i \textless LMS.length(); i++}
		\FOR{each layer in $LMS[i]$}
		\STATE $C_{LMS[i]}$ $\gets$ $C_{LMS[i]}$.concatenate(layer);
		\ENDFOR
		\ENDFOR
		
		\FOR{j = 0; j \textless LMS.length(); j++}
		\FOR{k = j+1; k \textless LMS.length(); k++}
		\STATE $\rho \gets  E[(C_{LMS[j]}-\mu_{C_{LMS[j]}})(C_{LMS[k]}-\mu_{C_{LMS[k]}})] /  \sigma_{C_{LMS[j]}}\sigma_{C_{LMS[k]}}$;
		\ENDFOR
		
		\STATE $PM[j][k]$ = $\rho$;
		$PM[k][j]$ $\gets$ $\rho$;
		\ENDFOR
		
		\textit{cluster\_result} $\gets$ spectral\_clustering($PM$,  $Num_{cluster}$ )
		
		\FOR{$\lambda$ = 0; $\lambda$ \textless $Num_{cluster}$; $\lambda$ ++}
		
		\FOR{$lm$ in \textit{cluster\_result}}
		
		\STATE $Sum_{lm} \gets Sum_{lm} + lm $ 
		
		\ENDFOR
		
		\STATE $CL_{a;\lambda}^{lm} \gets Sum_{lm} / Num_{cluster}$;			
		$An_{num} \gets 0 $;
		
		\FOR{$An$ in $AN$}
		\STATE $SUM_{acc} \gets SUM_{acc} + An.validate(CL_{a;\lambda}^{lm})$
		\STATE $An_{num} \gets An_{num} + 1$
		\ENDFOR 
		\STATE $AL$[$\lambda$] $\gets$ $SUM_{acc}/ An_{num}$
		\ENDFOR
		\STATE $Max\_list $ $\gets$ max($AL$);
		$AL$  $\gets$ $lm$ in $Max\_list$;
		\STATE  $\theta$ $\gets$ $LT$ which generates $lm$ in $Max\_list$
		\STATE $\nu$ = $AL$.length();
		$gm_{\epsilon}^{\theta} \gets \sum\limits_{\lambda = 1}^{\nu}  CL_{a;\lambda}^{lm}  / \nu$
		
		\ENSURE  $gm_{\epsilon}^{\theta}$ : Global Model in $\epsilon$ round
	\end{algorithmic}  
\end{algorithm}

\subsection{Block Package}

Storing models on the blockchain ensures data security. However, if all the local models generated by participating nodes in federated learning training are stored on the blockchain, it would impose significant storage pressure on the nodes. 

\begin{figure}[H]
	\centering
	\includegraphics[width=\columnwidth]{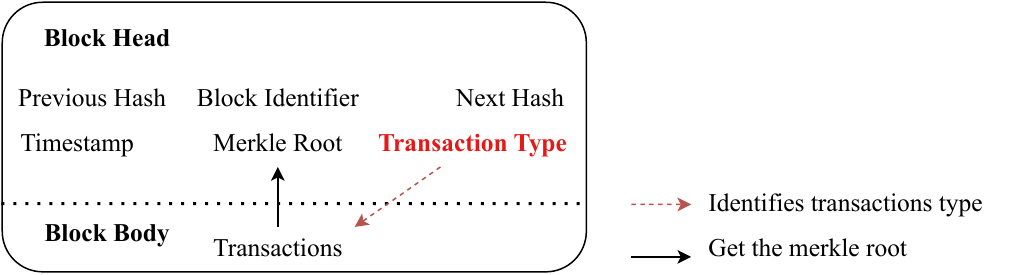}
	\caption{Block Structure in BRFL}
	\label{Block Structure}
\end{figure}

To address this problem, we propose a new transaction type (denoted as $\mathbb{T}$) called identity in the block header to distinguish different types of transactions in the block body, as shown in Fig. \ref{Block Structure}. This approach allows nodes in the blockchain to save specific blocks associated with a particular transaction type ($\mathbb{T}$), thereby reducing the storage resources required by the nodes.

Once the global model aggregation is completed, the Federated Blockchain Aggregation Manager (BRFL) packages the following components into distinct blocks: the Contract Identifier (CID), global model, accuracy records, and PCC transactions. The  $\mathbb{T}$ can be categorized into 4 types:

\begin{enumerate}[1.]
	
	\item Local Model Link CID ($\mathbb{T}_{CID}$): The transactions in the block body contain the Contract Identifier (CID) linked to the local model stored in IPFS. Transactions with  CID in a round will be packed into the same block.
	
	\item Global Model ($\mathbb{T}_{GM}$): The transactions in the block contains the global model.
	
	\item Accuracy Verification Record ($\mathbb{T}_{AVR}$): The transactions contains the accuracy record of $CL_{a;\lambda}^{lm}$ verified by $AN$.
	
	\item Pearson Correlation Coefficient Records ($\mathbb{T}_{PCC}$): The transactions contains the PCC records between the local model and the global model in the previous round, as well as the PCC records between the local model and the local model in the current round.
\end{enumerate}

The nodes in the blockchain have the flexibility to choose which type of transaction type ($\mathbb{T}$) blocks they want to save. In the case of $LT$, the $\mathbb{T}_{GM}$ block is necessary, while the $\mathbb{T}_{CID}$ block, $\mathbb{T}_{AVR}$ block, and $\mathbb{T}_{PCC}$ block are optional. On the other hand, for each round of $AN$, all four blocks ($\mathbb{T}_{GM}$, $\mathbb{T}_{CID}$, $\mathbb{T}_{AVR}$, and $\mathbb{T}_{PCC}$) must be downloaded to avoid any security issues arising from unsaved blocks. The $BN$ node has the capability to save all the blocks to maintain ledger security through consensus.
%
%
%
%

\section{PERFORMANCE EVALUATION}
\label{PERFORMANCE EVALUATION}

\subsection{Datasets}
\begin{itemize}
	
	\item MNIST: The MNIST dataset comprises handwritten numbers from 250 different individuals, totaling 70,000 images. The training set contains 60,000 images, while the test set contains 10,000 images. All images are grayscale and have dimensions of 28×28 pixels, with each image featuring a single handwritten digit.
	
	\item Fashion-MNIST: The Fashion-MNIST dataset consists of 70,000 frontal images of various fashion products across 10 categories. The dataset's size, format, and division into training and test sets are consistent with the MNIST dataset. It also follows the 60,000/10,000 division for training and testing, with images of 28×28 pixels in grayscale.
\end{itemize}

\subsection{Attacks}
\begin{enumerate}
	\item Noise attack: The noise  attacks put random noise variance  to the update gradients or local iterates.
	\item Sign-flipping Attack(SF) 	\cite{Li2018RSAB}: 
	The sign-flipping attacks flip the signs of update gradients or local iterates and enlarge the magnitudes.
\end{enumerate}

\subsection{Baselines}

Krum: 

The Krum  \cite{Blanchard2017MachineLW} aggregation algorithm selects the centroid of the local model parameters as the global model. It considers the Euclidean distance between a malicious gradient and a benign gradient to be large, while the Euclidean distance between two benign gradients is small.

Median: 
Median \cite{Yin2018ByzantineRobustDL}, proposed in 2021, aggregates input gradients by calculating the median of the values for each dimension of the local gradients. Assuming there are $N_{Median}$ local models, the algorithm collects and sorts the gradients in each dimension. For each dimension, the median gradient is selected as the gradient of the global model in that dimension.

Mean: Mean aggregates its input gradient by calculating the mean value of each dimension of local gradient.

Clippedclustering:  Clippedclustering \cite{Li2023AnES} proposed in 2023 , designed an automated clipping strategy to defend against potentially amplifying malicious updates. 

\begin{table}[htbp]
	\centering
	\caption{Model Parameters}
	\begin{tabularx}{\columnwidth}{Xr}
		\toprule
		parameter & value \\
		\midrule
		Batch size of client's local training& 10  \\
		Optimizer  of client's local training& SGD  \\
		Learning Rate  of client's local training& 0.01  \\
		Local Epochs  of client's local training& 5  \\
		Max running round  of client's local training& 100 \\
		Noise Variance(in noise attack) & 1  \\
		Numer of Spectral Clusters (in PSA) & 2  \\
		Number of aggregation nodes  (in PPCC) & 2 \\
		Number of local training nodes & 10 \\
		Number of blockchain network nodes & 2 \\
		Initial stake for blockchain nodes& 0 \\
		The fixed number of  stake benefit for client's local training in one round (FS)& 20 \\
		Stake benefit for aggregation node in one round & 2 \\
		\bottomrule
	\end{tabularx}
	\end{table}
	
	\subsection{Accuracy }
	\label{Accuracy}
	In this section, we evaluate the performance of PSA in both noise attack and sign-flipping attack scenarios. We compare the performance of PSA with other aggregation algorithms, namely Krum, Median, Mean, and Clipped Clustering, in the context of BRFL.
	
	Noise attack and sign-flipping attack are types of model poison attacks, where malicious clients modify model gradients after local training. In Figure \ref{Accuracy}, we present the results of testing these five different aggregation algorithms under noise attack and sign-flipping attack scenarios. The evaluations are conducted on the MNIST and Fashion-MNIST datasets, considering both independent and identically distributed (IID) data distribution and non-IID data distribution. Additionally, we set the number of malicious clients to be half of the total number of local training clients.
	
\begin{figure*}[htbp]
\centering

\begin{subfigure}[b]{0.22\textwidth}
	\centering
	\includegraphics[width=\textwidth]{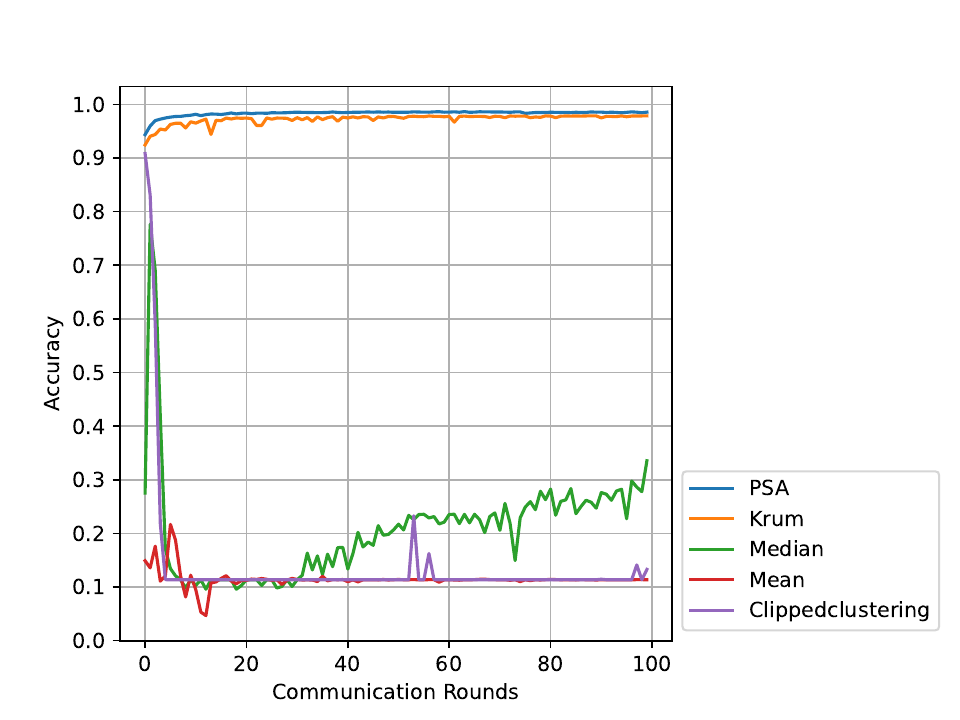}
	\caption{MNIST-IID-Noise}
	\label{mnist-iid-noise-50-acc}
\end{subfigure} 
\begin{subfigure}[b]{0.22\textwidth}
	\centering
	\includegraphics[width=\textwidth]{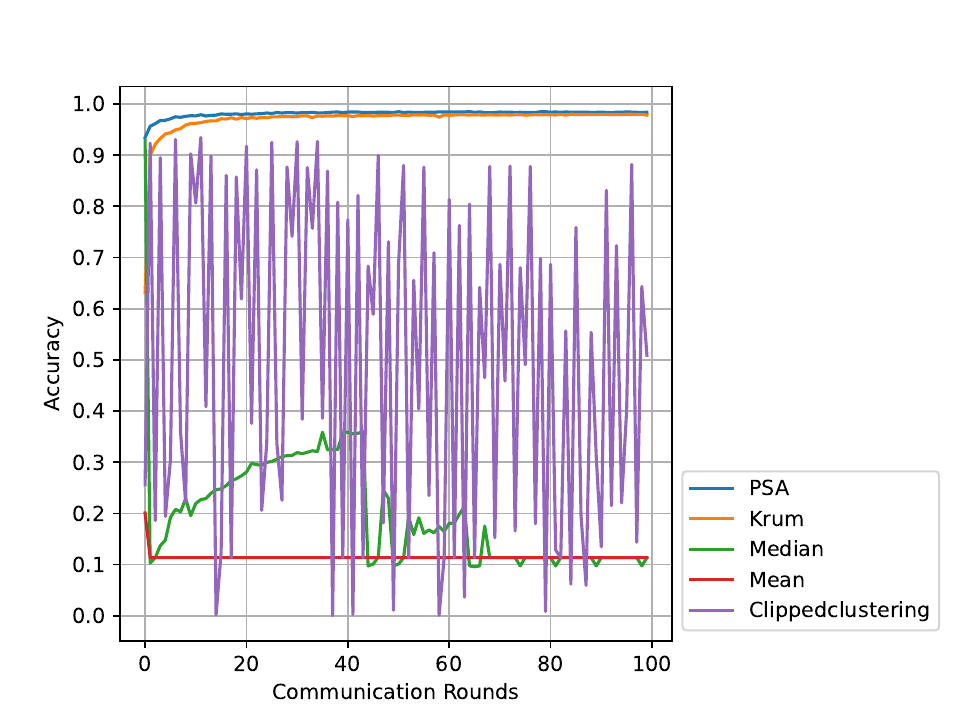}
	\caption{MNIST-IID-Signflip}
	\label{mnist-iid-signflip-50-acc}
\end{subfigure}
\begin{subfigure}[b]{0.22\textwidth}
	\centering
	\includegraphics[width=\textwidth]{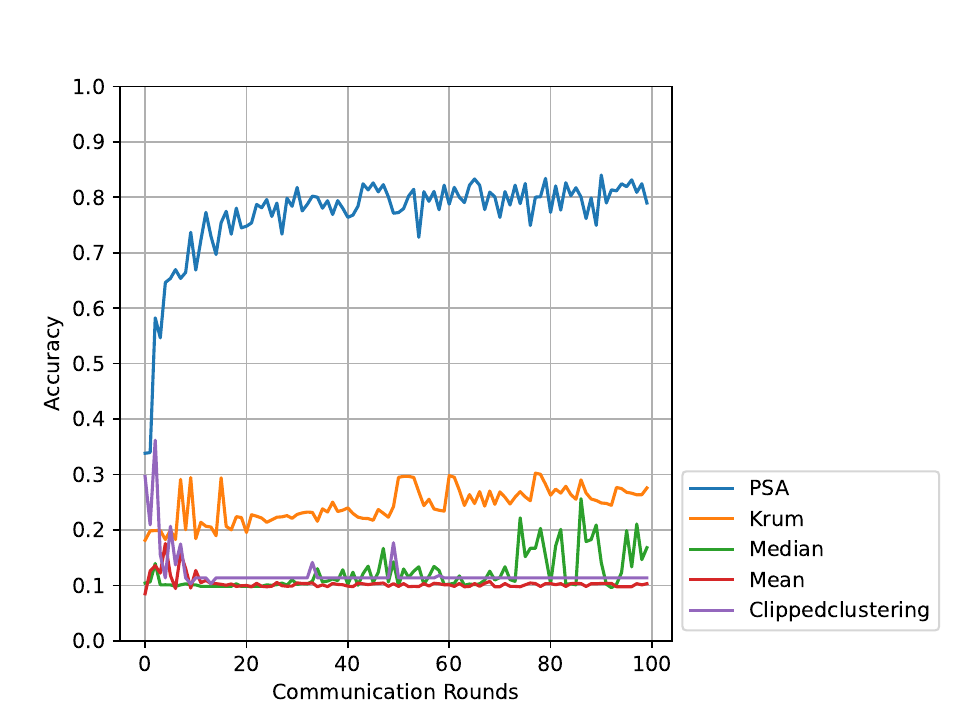}
	\caption{MNIST-Non-IID-Noise}
	\label{mnist-non-iid-noise-50-acc}
\end{subfigure}
\begin{subfigure}[b]{0.22\textwidth}
	\centering
	\includegraphics[width=\textwidth]{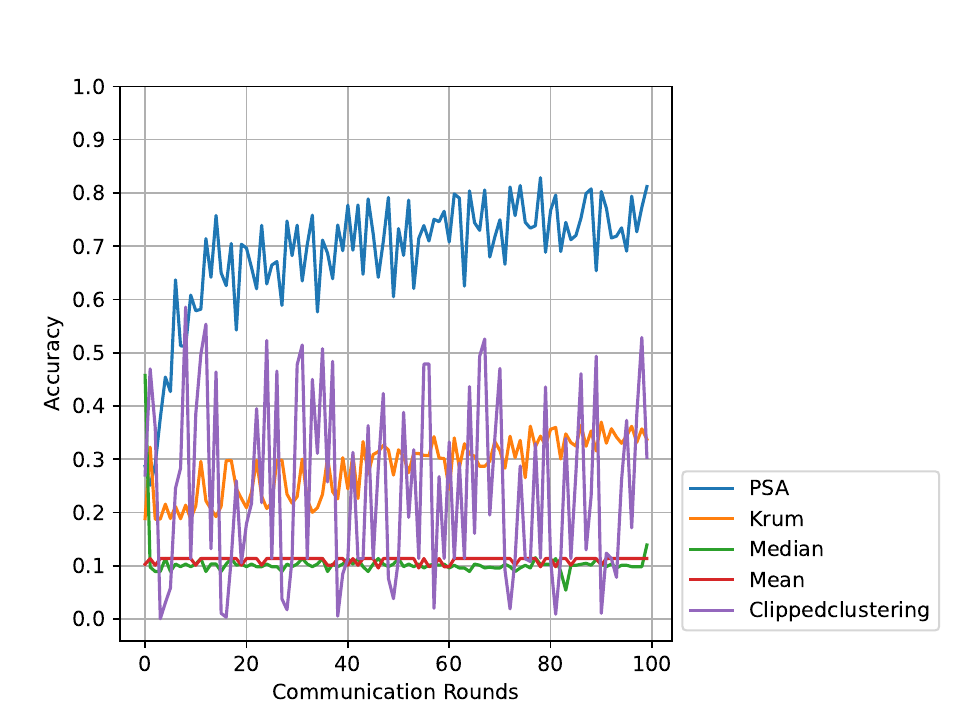}
	\caption{MNIST-Non-IID-Signflip}
	\label{mnist-non-iid-signflip-50-acc}
\end{subfigure}

\begin{subfigure}[b]{0.22\textwidth}
	\centering
	\includegraphics[width=\textwidth]{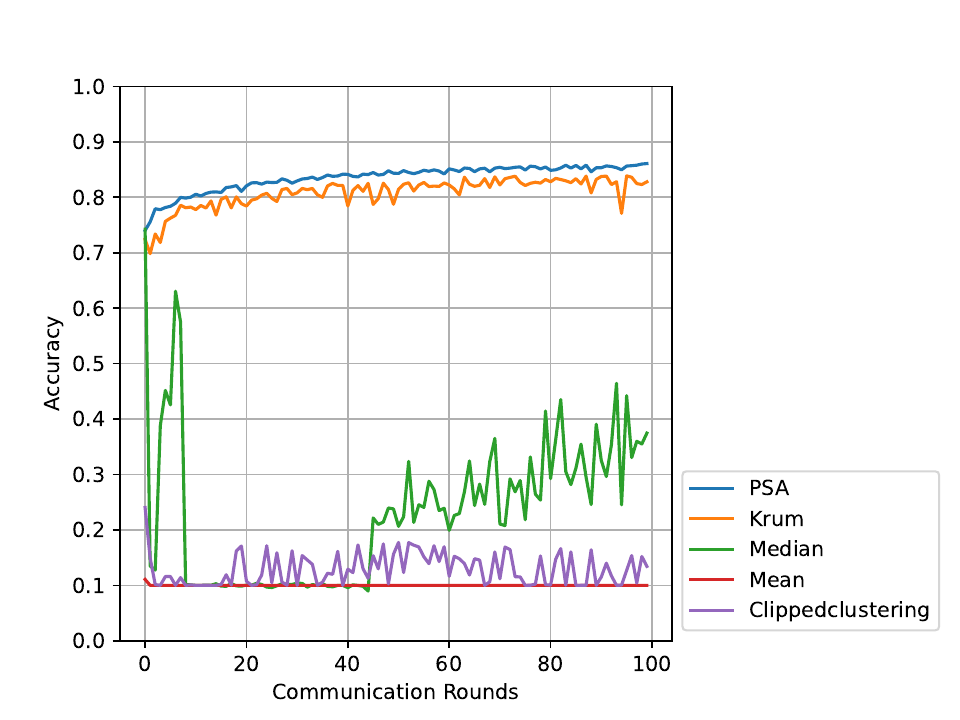}
	\caption{Fashion-MNIST-IID-Noise}
	\label{fashionmnist-iid-noise-50-acc}
\end{subfigure}
\begin{subfigure}[b]{0.22\textwidth}
	\centering
	\includegraphics[width=\textwidth]{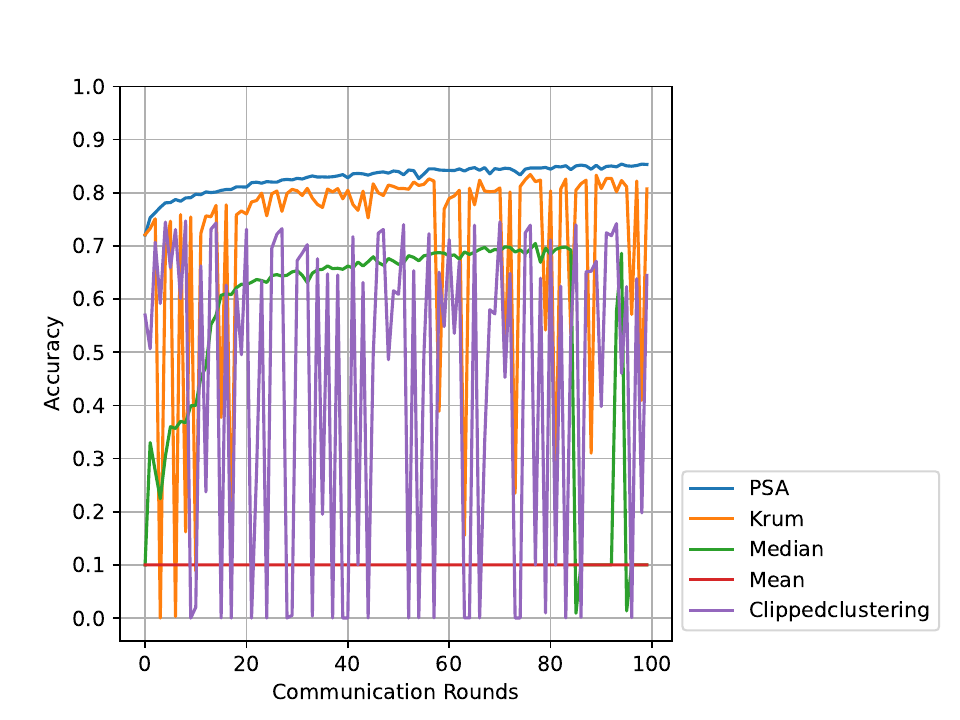}
	\caption{Fashion-MNIST-IID-Signflip}
	\label{fashionmnist-iid-signflip-50-acc}
\end{subfigure}	
\begin{subfigure}[b]{0.22\textwidth}
	\centering
	\includegraphics[width=\textwidth]{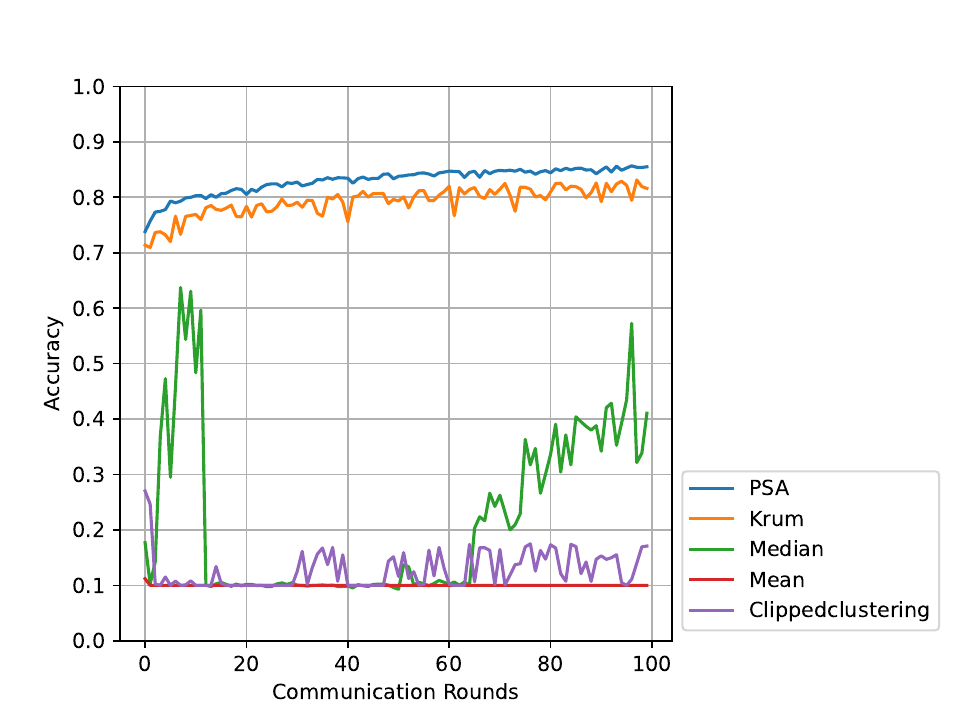}
	\caption{Fashion-MNIST-Non-IID-Noise}
	\label{fashionmnist-non-iid-noise-50-acc}
\end{subfigure}
\begin{subfigure}[b]{0.22\textwidth}
	\centering
	\includegraphics[width=\textwidth]{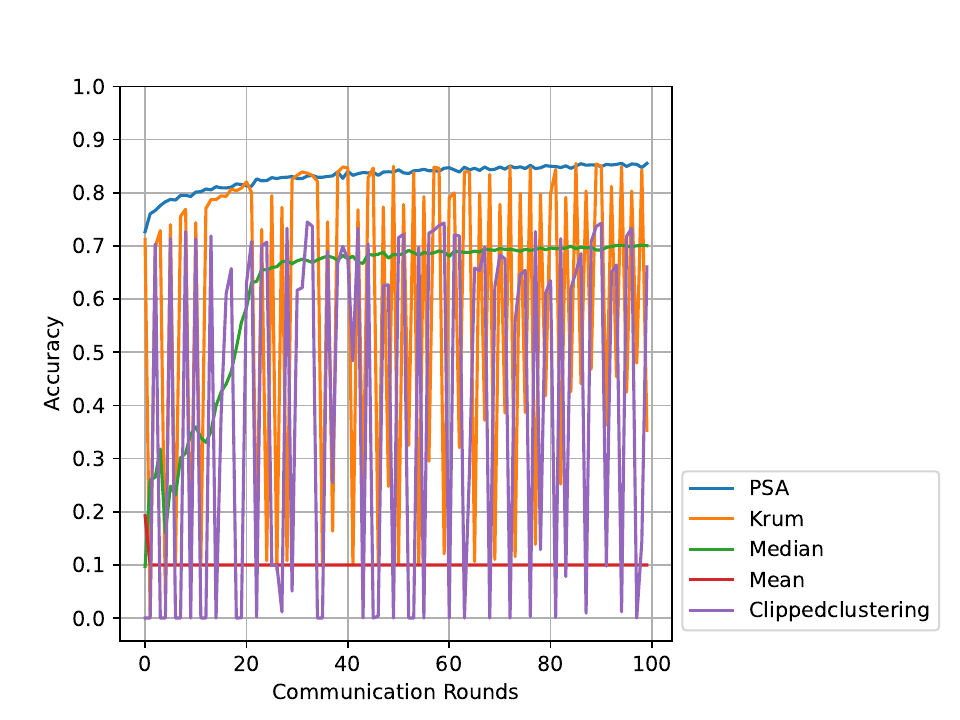}
	\caption{Fashion-MNIST-Non-IID-Signflip}
	\label{fashionmnist-non-iid-signflip-50-acc}
\end{subfigure}

\caption{Accuracy of BRLF, Krum, Median, Mean, Clippedclustering under 50\% byzantine attack for MNIST and Fashion-MNIST task }
\label{Accuracy}
\end{figure*}

	Figure \ref{mnist-iid-noise-50-acc} illustrates the accuracy comparison between our proposed PSA and the baselines when using the MNIST dataset with IID data distribution and noise attack as the model poison attack. From Figure \ref{mnist-iid-noise-50-acc}, we observe that our PSA achieves the highest accuracy among all the algorithms. Krum, median, mean, and clipped clustering exhibit lower performance when 50\% of the clients are malicious. Similarly, in Figure \ref{mnist-iid-signflip-50-acc}, which represents the performance under sign-flipping attacks on the MNIST dataset with IID data distribution, PSA and Krum outperform the other baselines. However, when the dataset distribution changes from IID to Non-IID, PSA maintains a consistently high performance, while Krum's accuracy decreases compared to PSA. PSA demonstrates superior performance in the Non-IID data distribution. In Figure \ref{mnist-non-iid-noise-50-acc}, Krum, median, mean, and clipped clustering exhibit stable performance, but PSA outperforms them all. Figure \ref{mnist-non-iid-signflip-50-acc} shows that clipped clustering exhibits large fluctuations, remaining in the lower range. Apart from PSA, the accuracy of the four baselines is below 0.5.
	
	When using the Fashion-MNIST dataset, as shown in Figure \ref{fashionmnist-iid-noise-50-acc}, PSA achieves the best accuracy when the data distribution is IID and the attack type is noise. Krum performs slightly worse than PSA but still above 0.8. The accuracy of the other three baselines is below 0.5. In Figure \ref{fashionmnist-iid-signflip-50-acc}, with the attack type changing to sign-flipping compared to Figure \ref{fashionmnist-iid-noise-50-acc}, PSA maintains stable and high performance. Krum and clipped clustering exhibit large fluctuations, while median remains stable for the first 80 rounds of training but starts to fluctuate later. The mean aggregation algorithm performs poorly. When the data distribution is Non-IID, as shown in Figures \ref{fashionmnist-non-iid-noise-50-acc} and \ref{fashionmnist-non-iid-signflip-50-acc}, PSA outperforms Krum and the other three baselines in Figure \ref{fashionmnist-non-iid-noise-50-acc} under noise attack. In Figure \ref{fashionmnist-non-iid-signflip-50-acc}, with sign-flipping attack as the attack type, PSA maintains high and stable performance, while Krum and clipped clustering exhibit large fluctuations and fail to surpass PSA. Median achieves an accuracy of 0.7, whereas the mean algorithm remains at 0.1.

\subsection{Stake Trends }
\label{Stake Trends}

In this section, we discuss the stake trends of different types of nodes in BRFL under various datasets, data distributions, and attack types.

\begin{figure*}[htbp]
\centering

\begin{subfigure}[b]{0.22\textwidth}
	\centering
	\includegraphics[width=\textwidth]{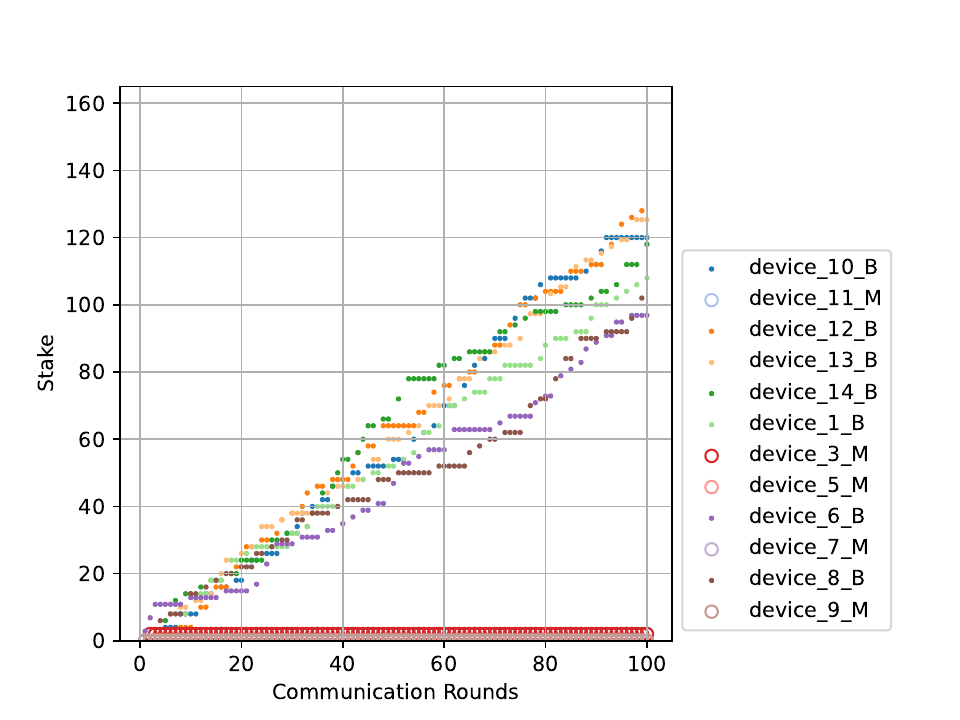}
	\caption{MNIST-IID-Noise}
	\label{mnist-iid-noise-50-stake_plots}
\end{subfigure}
\begin{subfigure}[b]{0.22\textwidth}
	\centering
	\includegraphics[width=\textwidth]{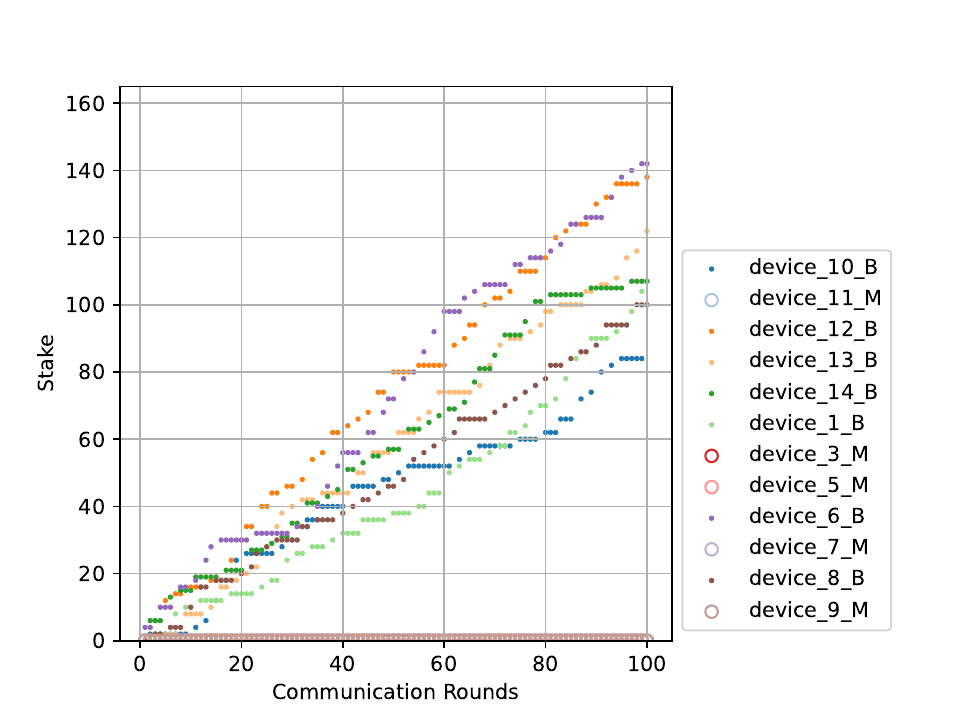}
	\caption{MNIST-IID-Signflip}
	\label{mnist-iid-signflip-50-stake_plots}
\end{subfigure}
\begin{subfigure}[b]{0.22\textwidth}
	\centering
	\includegraphics[width=\textwidth]{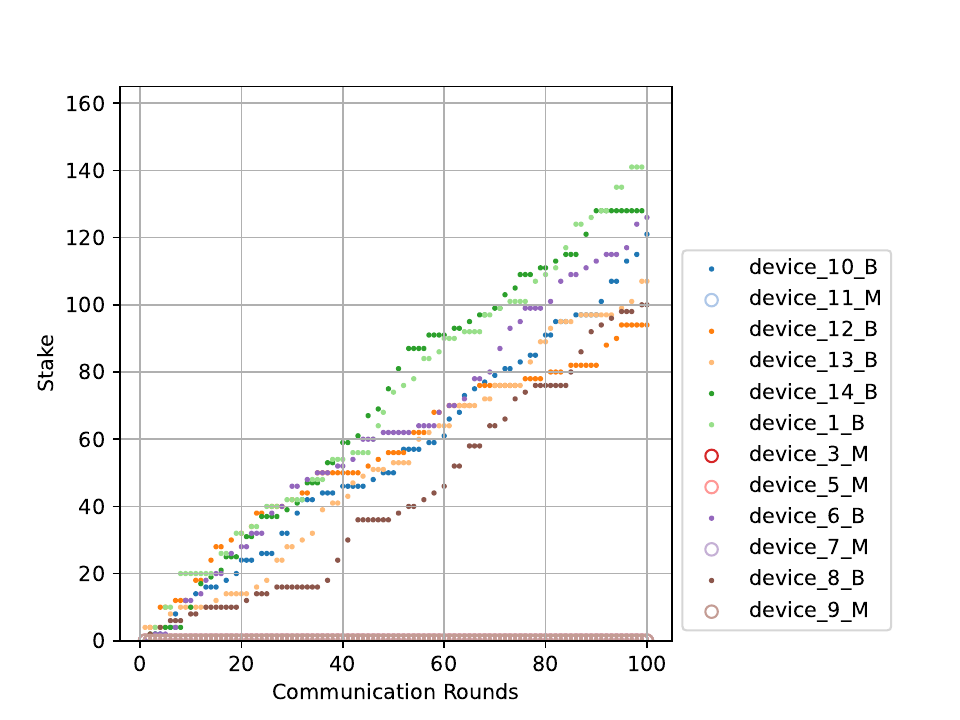}
	\caption{MNIST-Non-IID-Noise}
	\label{mnist-non-iid-noise-50-stake_plots}
\end{subfigure}
\begin{subfigure}[b]{0.22\textwidth}
	\centering
	\includegraphics[width=\textwidth]{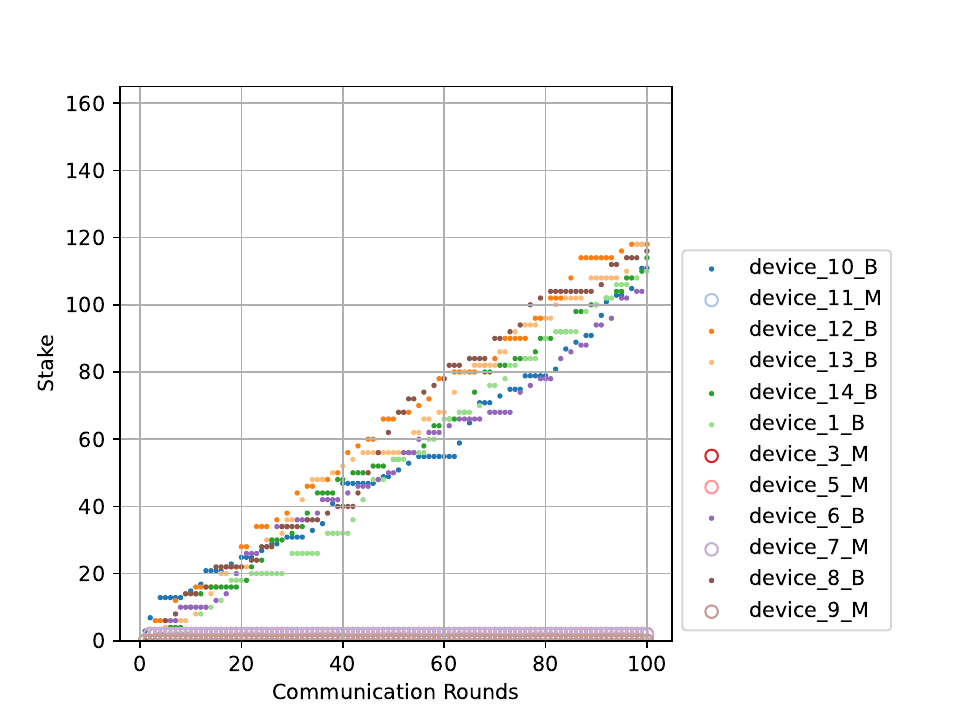}
	\caption{MNIST-Non-IID-Signflip}
	\label{mnist-non-iid-signflip-50-stake_plots}
\end{subfigure}

\begin{subfigure}[b]{0.22\textwidth}
	\centering
	\includegraphics[width=\textwidth]{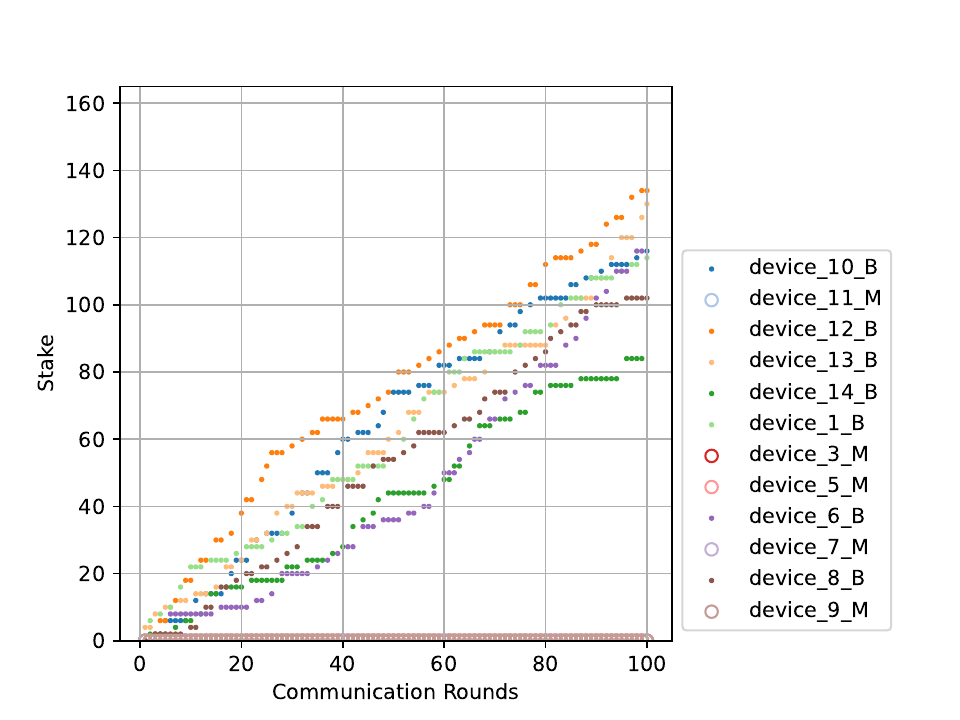}
	\caption{Fashion-MNIST-IID-Noise}
	\label{fashionmnist-iid-noise-50-stake_plots}
\end{subfigure}
\begin{subfigure}[b]{0.22\textwidth}
	\centering
	\includegraphics[width=\textwidth]{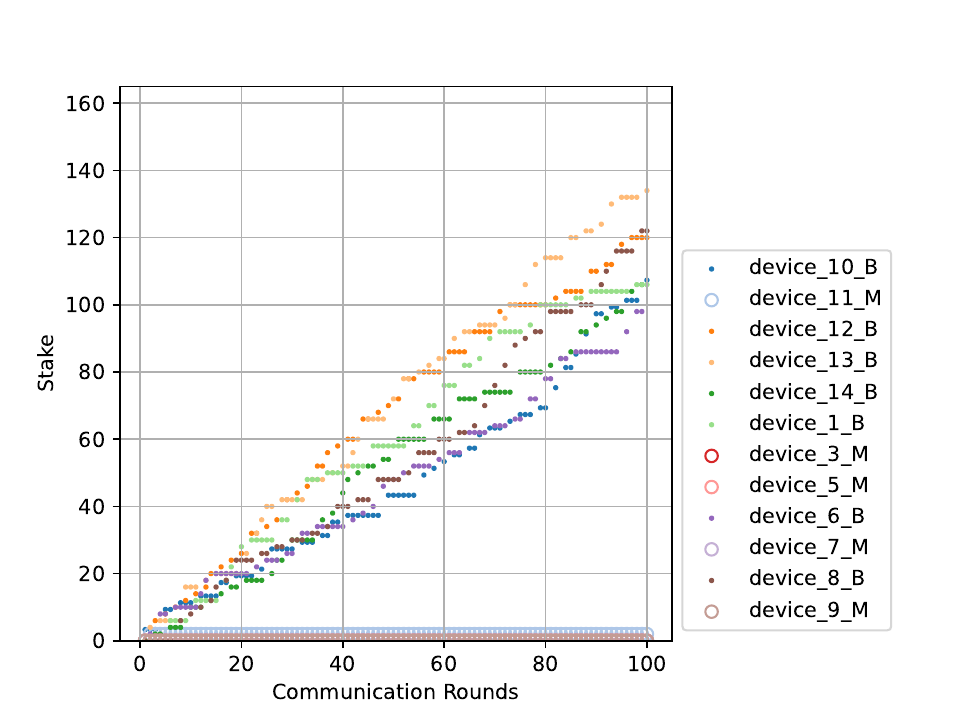}
	\caption{Fashion-MNIST-IID-Signflip}
	\label{fashionmnist-iid-signflip-50-stake_plots}
\end{subfigure}
\begin{subfigure}[b]{0.22\textwidth}
	\centering
	\includegraphics[width=\textwidth]{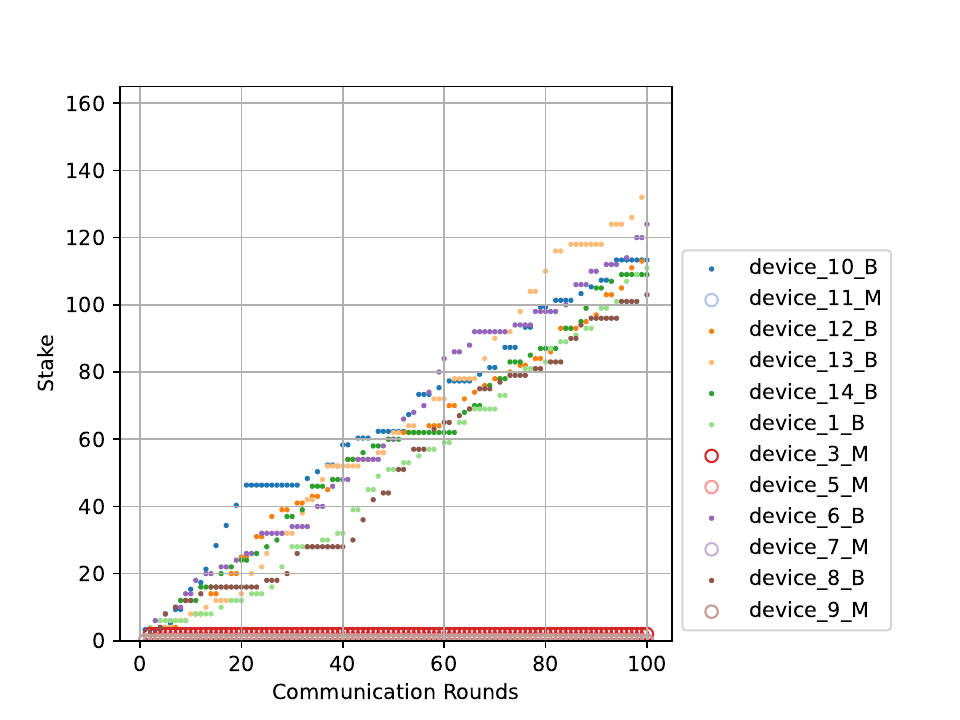}
	\caption{Fashion-MNIST-Non-IID-Noise}
	\label{fashionmnist-non-iid-noise-50-stake_plots}
\end{subfigure}
\begin{subfigure}[b]{0.22\textwidth}
	\centering
	\includegraphics[width=\textwidth]{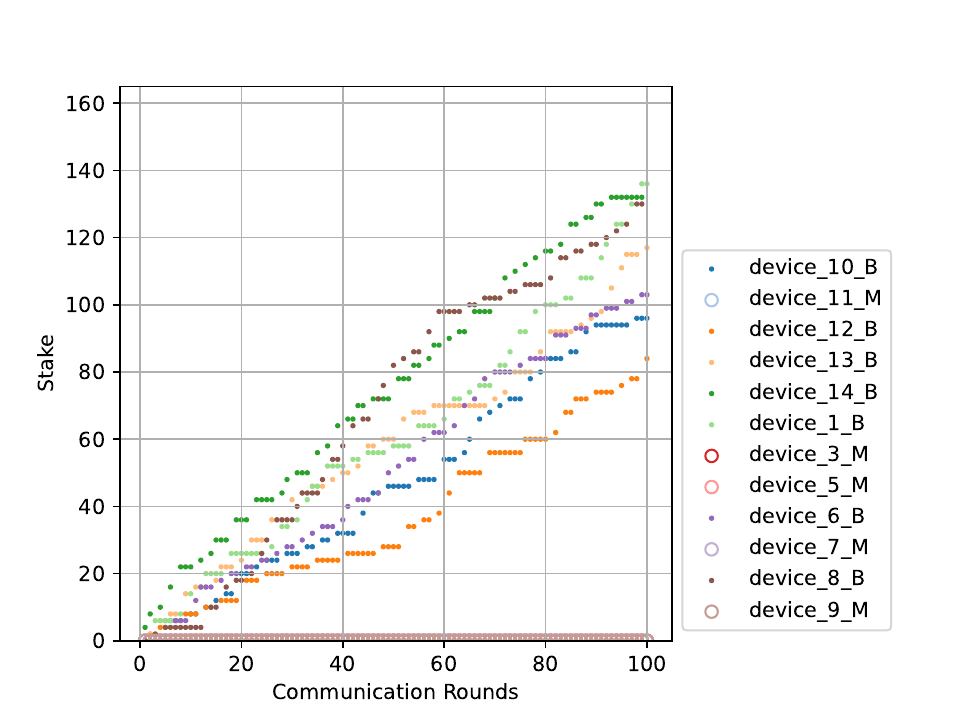}
	\caption{Fashion-MNIST-Non-IID-Signflip}
	\label{fashionmnist-non-iid-signflip-50-stake_plots}
\end{subfigure}

\caption{Stake  of BRLF under 50\% byzantine attack for MNIST and Fashion-MNIST task }
\label{Stake}
\end{figure*}

The labels of Figures \ref{Stake} consist of two parts: the client's name and whether the client is malicious. The clients are named from \textit{device\_1} to \textit{device\_14}, except for \textit{device\_2} and \textit{device\_4}, which are blockchain network nodes. Thus, we did not count their stakes. The letter \textit{B} after the client's name indicates a benign client, while \textit{M} signifies a malicious client. We use dots to represent the stakes of benign clients and circles to represent the stakes of malicious clients. From Figures \ref{Stake}, we observe that the stakes of benign clients show a nonlinear increasing trend, while the stakes of malicious nodes remain stable around 0.

From Figure \ref{mnist-iid-noise-50-stake_plots}, we can see that the benign client named \textit{device\_12} has the largest number of stakes in the final result. This suggests that it either acts as an aggregation node or its local model ($lm$) is frequently selected for the $\mathbb{MAQ}$, resulting in the largest stake benefit from the consensus algorithm due to its effective benign local training work. The client named \textit{device\_13} has the second-largest number of stakes. In the last three rounds, its stake remains unchanged, indicating that although it participates in local model training, its $lm$ does not enter the $\mathbb{MAQ}$, and it does not become an aggregation node in these three rounds, resulting in its stake maintaining a constant value. Clients \textit{device\_11}, \textit{device\_3}, \textit{device\_5}, \textit{device\_7}, and \textit{device\_9} are malicious clients, and their stakes are around 0, indicating that their generated $lm$ rarely or never enters the $\mathbb{MAQ}$, which validates the effectiveness of our proposed BRFL. The figures in Figure \ref{Stake} correspond to Figure \ref{Accuracy} and depict the stake trends under different datasets, data distributions, and attack types when malicious clients account for 50\% of the total, which corresponds to 5 out of 10 local training nodes.

\subsection{Blockchain Size Trends }

\begin{figure*}[htbp]
\centering

\begin{subfigure}[b]{0.22\textwidth}
	\centering
	\includegraphics[width=\textwidth]{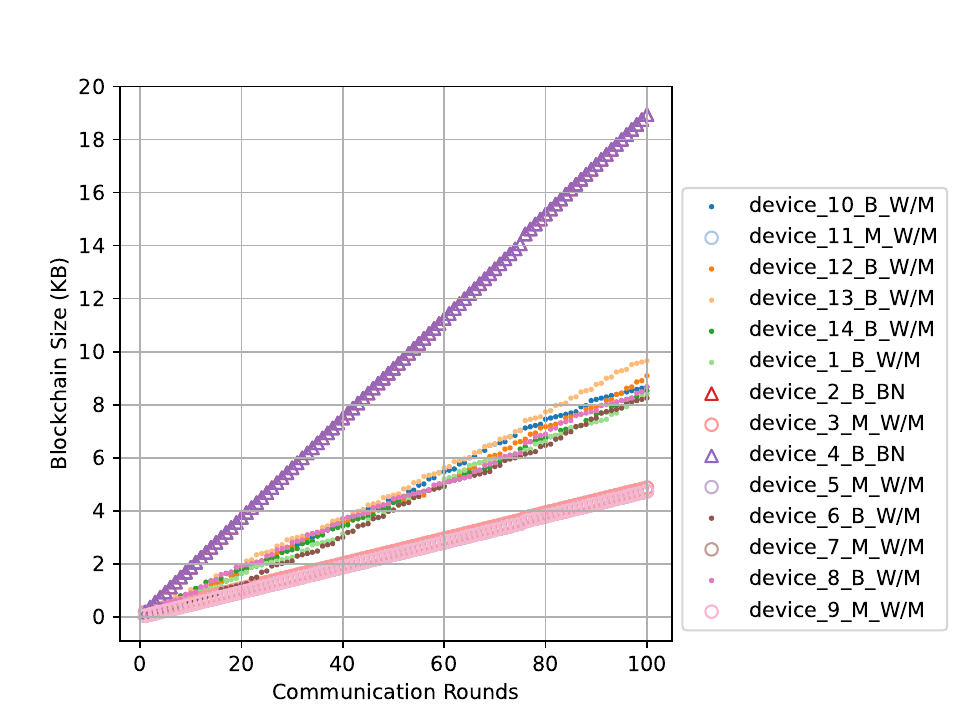}
	\caption{MNIST-IID-Noise}
	\label{mnist-iid-noise-50}
\end{subfigure}
\begin{subfigure}[b]{0.22\textwidth}
	\centering
	\includegraphics[width=\textwidth]{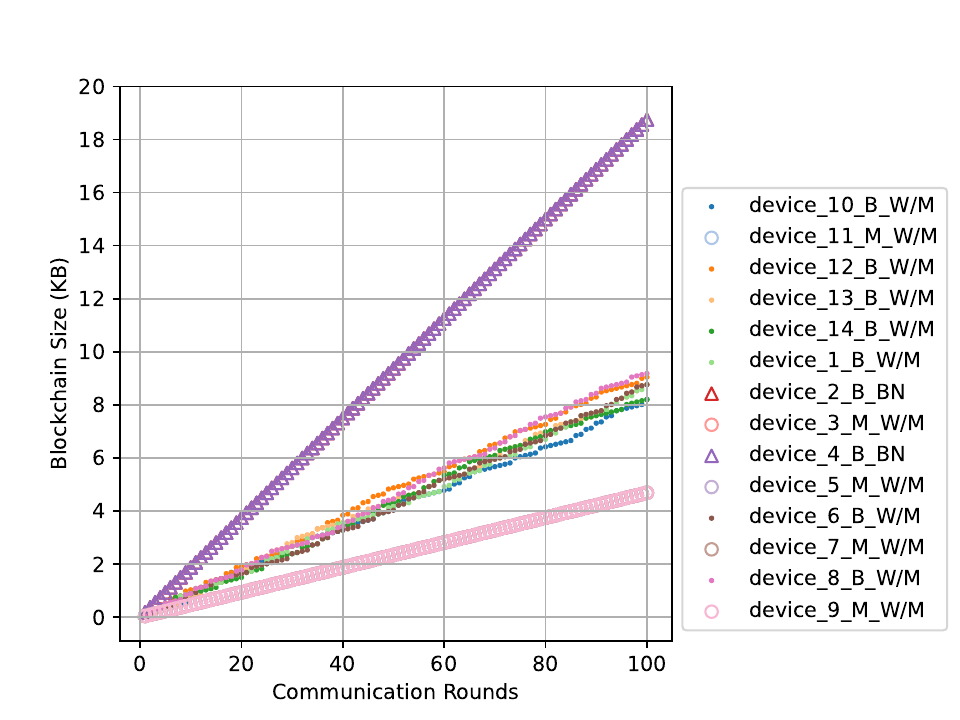}
	\caption{MNIST-IID-Signflip}
	\label{mnist-iid-signflip-50}
\end{subfigure}
\begin{subfigure}[b]{0.22\textwidth}
	\centering
	\includegraphics[width=\textwidth]{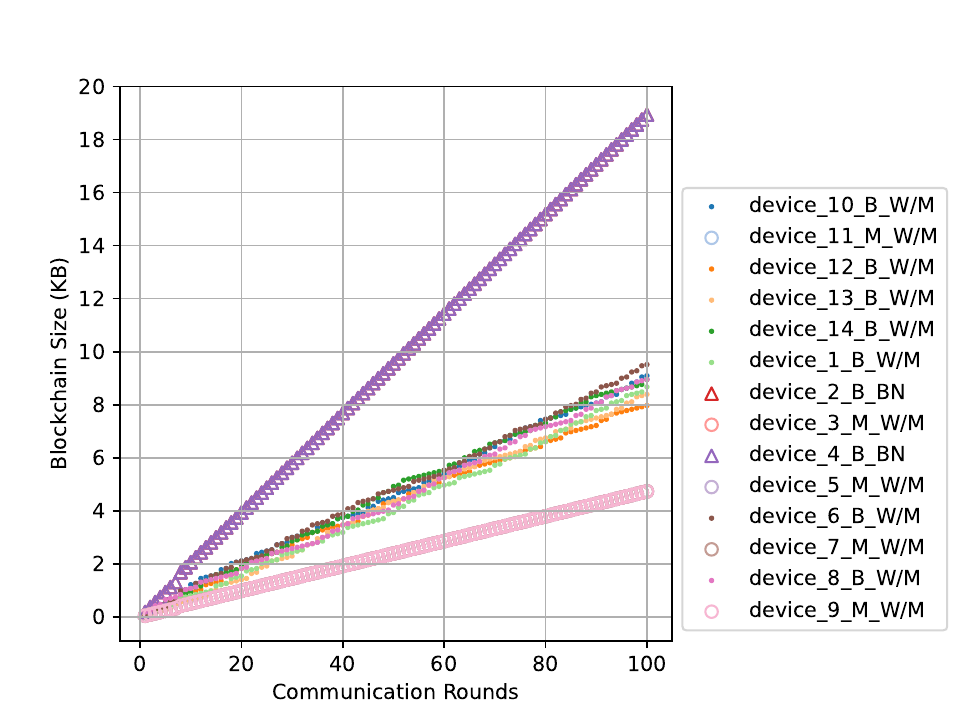}
	\caption{MNIST-Non-IID-Noise}
	\label{mnist-non-iid-noise-50}
\end{subfigure}
\begin{subfigure}[b]{0.22\textwidth}
	\centering
	\includegraphics[width=\textwidth]{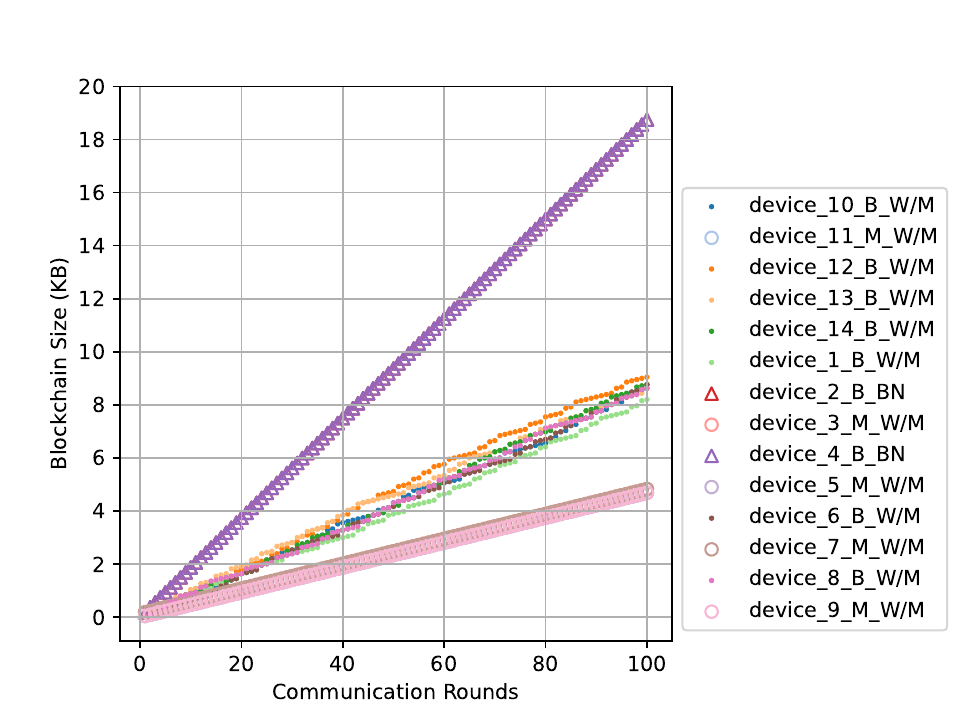}
	\caption{MNIST-Non-IID-Signflip}
	\label{mnist-non-iid-signflip-50}
\end{subfigure}

\begin{subfigure}[b]{0.22\textwidth}
	\centering
	\includegraphics[width=\textwidth]{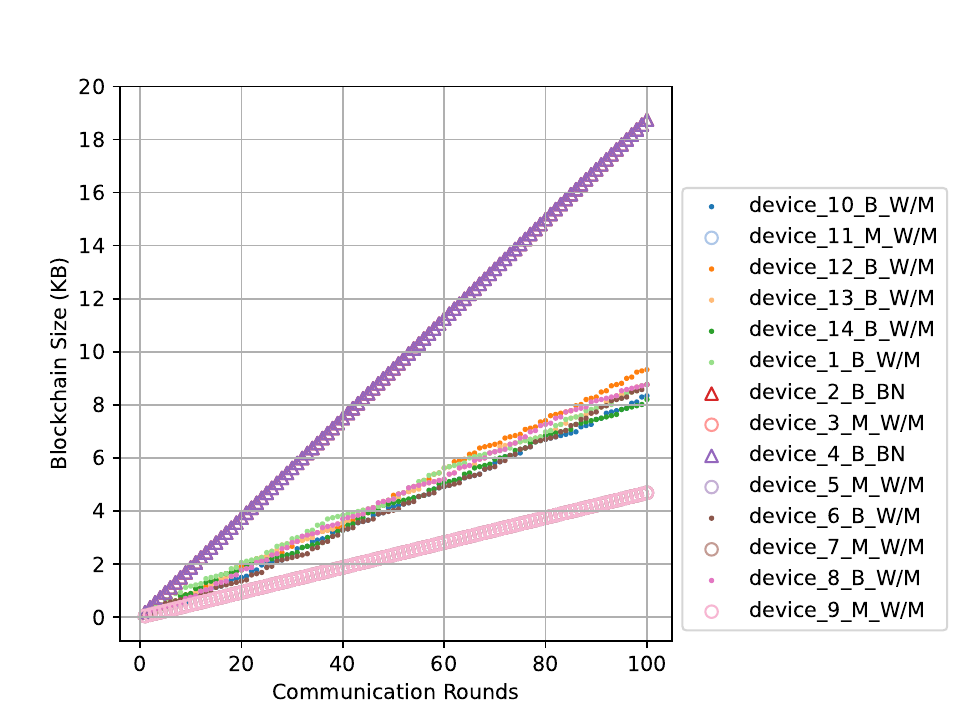}
	\caption{Fashion-MNIST-IID-Noise}
	\label{fashionmnist-iid-noise-50}
\end{subfigure}
\begin{subfigure}[b]{0.22\textwidth}
	\centering
	\includegraphics[width=\textwidth]{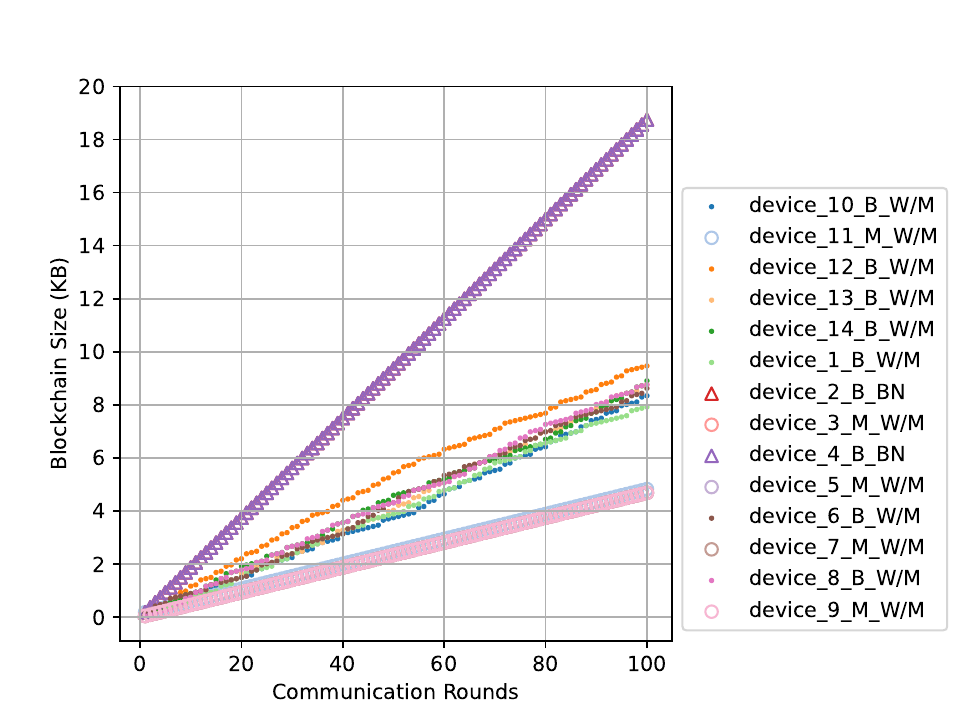}
	\caption{Fashion-MNIST-IID-Signflip}
	\label{fashionmnist-iid-signflip-50}
\end{subfigure}
\begin{subfigure}[b]{0.22\textwidth}
	\centering
	\includegraphics[width=\textwidth]{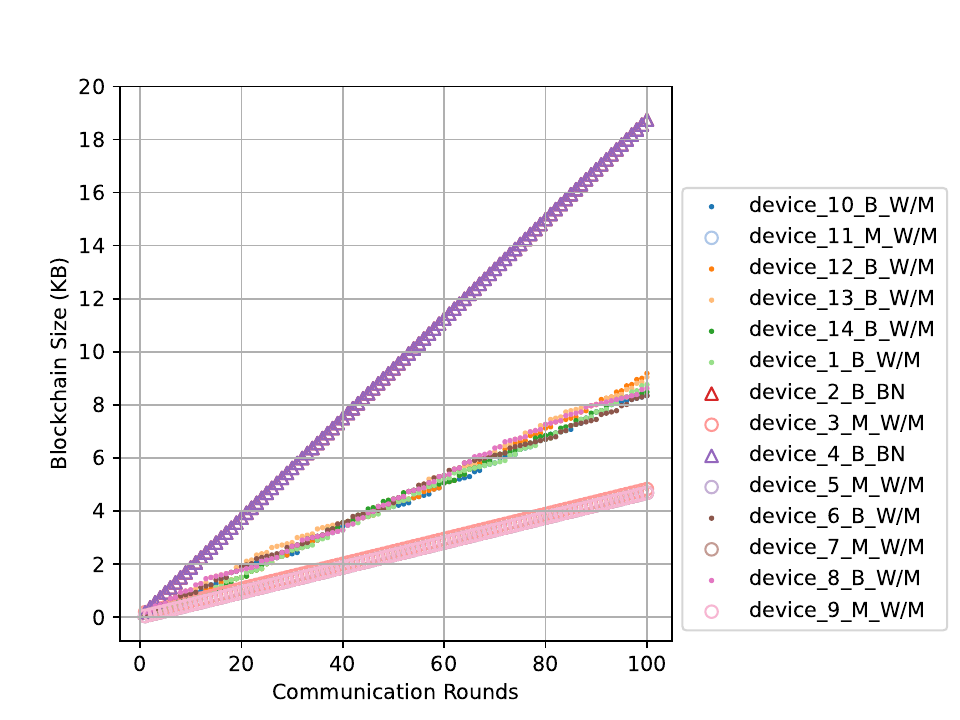}
	\caption{Fashion-MNIST-Non-IID-Noise}
	\label{fashionmnist-non-iid-noise-50}
\end{subfigure}
\begin{subfigure}[b]{0.22\textwidth}
	\centering
	\includegraphics[width=\textwidth]{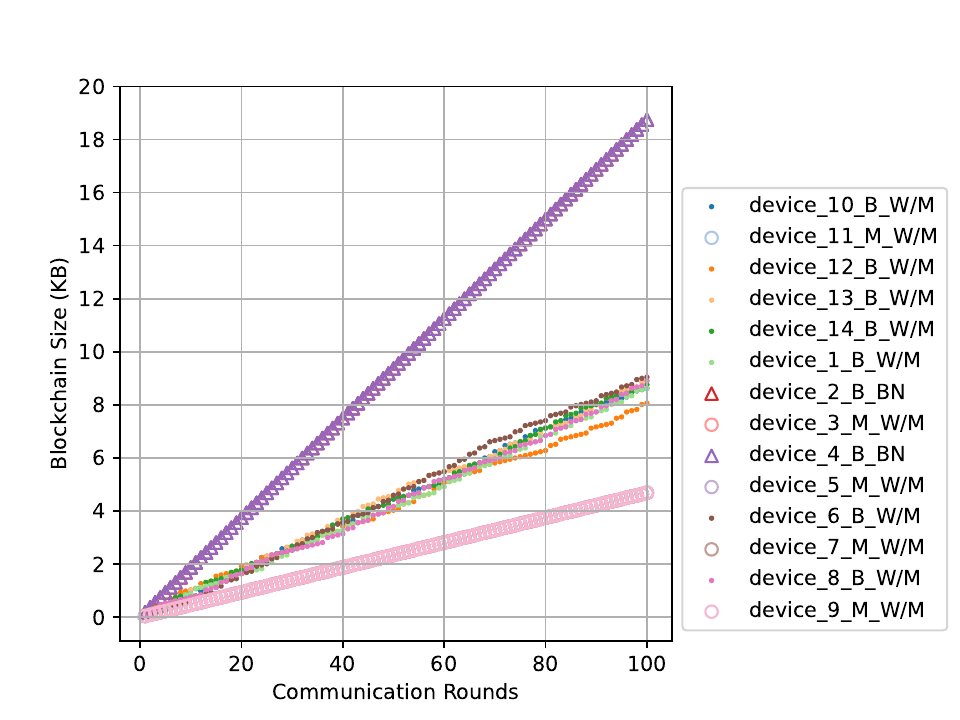}
	\caption{Fashion-MNIST-Non-IID-Signflip}
	\label{fashionmnist-non-iid-signflip-50}
\end{subfigure}

\caption{Blockchain size of BRLF under 50\% attack rate with two types of \textit{byzantine-attack} for MNIST and Fashion-MNIST tasks}
\label{blockchain_size}
\end{figure*}

In this section, we discuss the trend of blockchain size in clients within BRFL in the same environment as depicted in Figure \ref{Accuracy} and Figure \ref{Stake}. In Figure \ref{blockchain_size}, we examine the variation in the size of the blockchain footprint stored on different devices.

We use triangles to represent blockchain network nodes. In BRFL, both blockchain network nodes and aggregation nodes store all types of blocks, while local training nodes only need to store blocks of the global model type. Different colored dots represent benign nodes, while different colored circles represent malicious nodes. The label in Figure \ref{blockchain_size} consists of three parts: device name, whether it is a malicious node, and the node type. The device names range from \textit{device\_1} to \textit{device\_14}, with \textit{B} denoting a benign node, \textit{M} denoting a malicious node, \textit{W/M} denoting a node that can be either a local training node or an aggregation node, and \textit{BN} indicating that the node is a blockchain network node that does not participate in the training and aggregation of federated learning and only exists in the blockchain network.

Figure \ref{blockchain_size} reveals that the blockchain size stored by the blockchain network nodes exhibits a linear growth trend. This is because the blockchain network nodes store all types of blocks in the blockchain, and each round of the federated learning process produces blocks that are of the same or similar size. Among the nodes of type \textit{W/M}, the block size of the benign nodes shows a nonlinear increasing trend. This is because when a benign node becomes an aggregation node, it stores all types of blocks, whereas as a local training node, it only stores one type of block. As the benign node alternates between being an aggregation node and a local training node, the size of the blockchain it stores shows a nonlinear increasing trend. The blockchain size in malicious nodes also shows a linear increasing trend, but the increase is smaller. This is because in the BRFL model, malicious models are detected, and malicious nodes are prevented from becoming aggregation nodes. Therefore, malicious nodes only store blocks of the global model type as local training nodes.

\subsection{Number of Being Aggregation nodes Times Trends }

\begin{figure*}[htbp]
\centering

\begin{subfigure}[b]{0.22\textwidth}
	\centering
	\includegraphics[width=\textwidth]{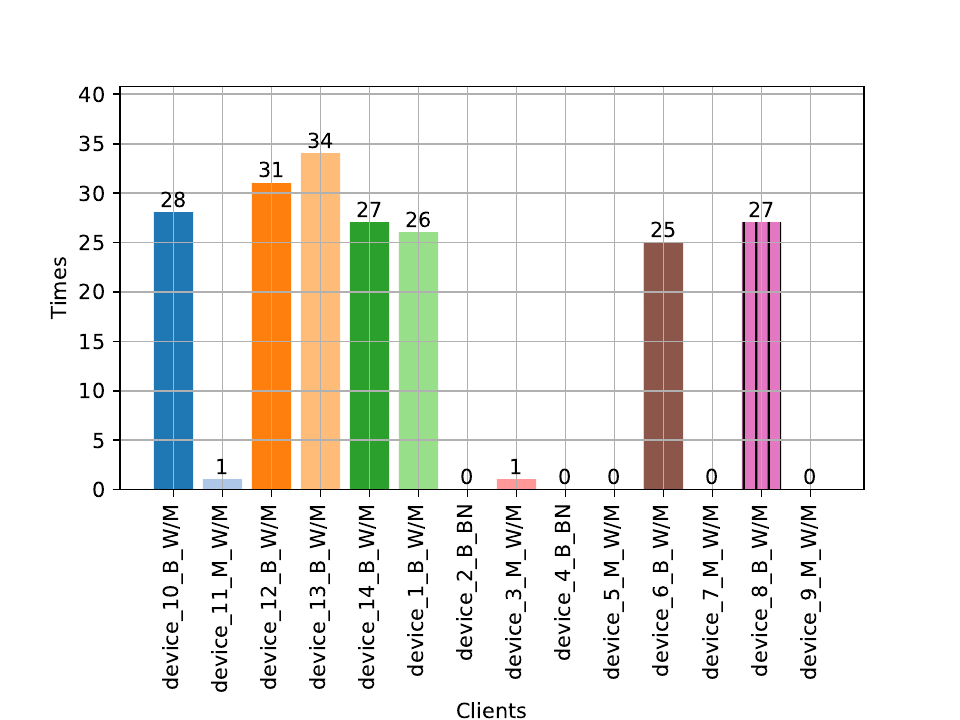}
	\caption{MNIST-IID-Noise}
	\label{mnist-iid-noise-50}
\end{subfigure}
\begin{subfigure}[b]{0.22\textwidth}
	\centering
	\includegraphics[width=\textwidth]{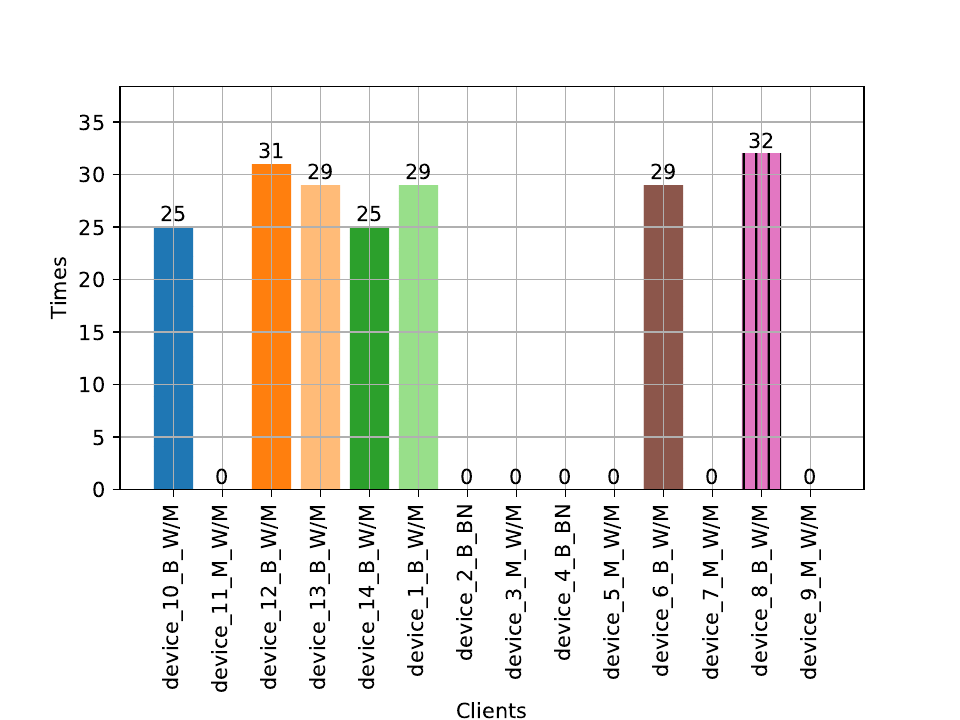}
	\caption{MNIST-IID-Signflip}
	\label{mnist-iid-signflip-50}
\end{subfigure}
\begin{subfigure}[b]{0.22\textwidth}
	\centering
	\includegraphics[width=\textwidth]{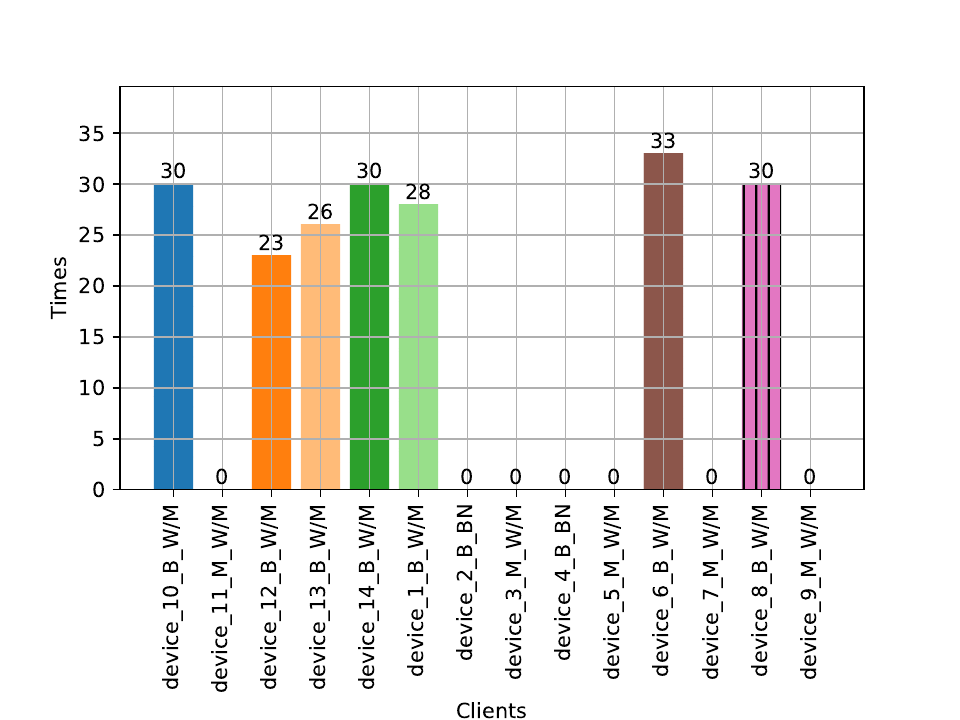}
	\caption{MNIST-Non-IID-Noise}
	\label{mnist-non-iid-noise-50}
\end{subfigure}
\begin{subfigure}[b]{0.22\textwidth}
	\centering
	\includegraphics[width=\textwidth]{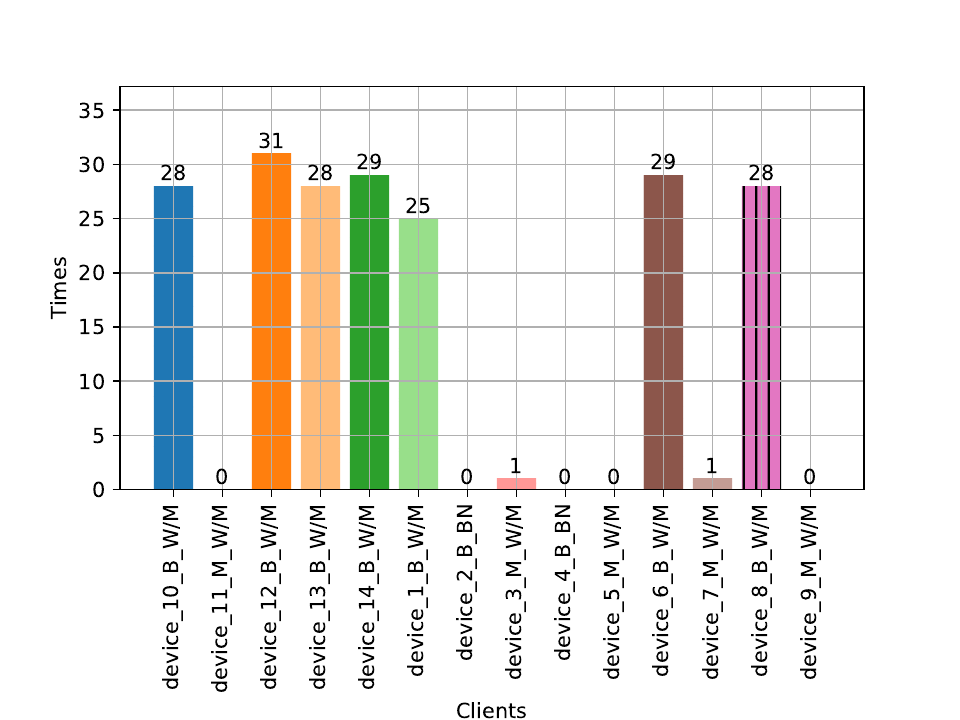}
	\caption{MNIST-Non-IID-Signflip}
	\label{mnist-non-iid-signflip-50}
\end{subfigure}

\begin{subfigure}[b]{0.22\textwidth}
	\centering
	\includegraphics[width=\textwidth]{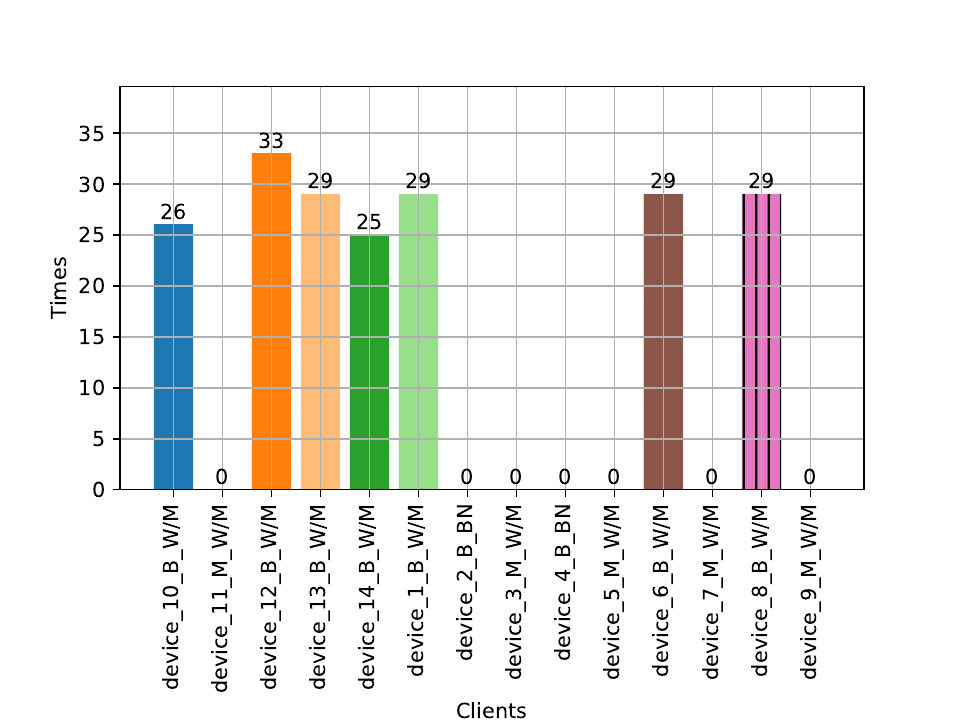}
	\caption{Fashion-MNIST-IID-Noise}
	\label{fashionmnist-iid-noise-50}
\end{subfigure}
\begin{subfigure}[b]{0.22\textwidth}
	\centering
	\includegraphics[width=\textwidth]{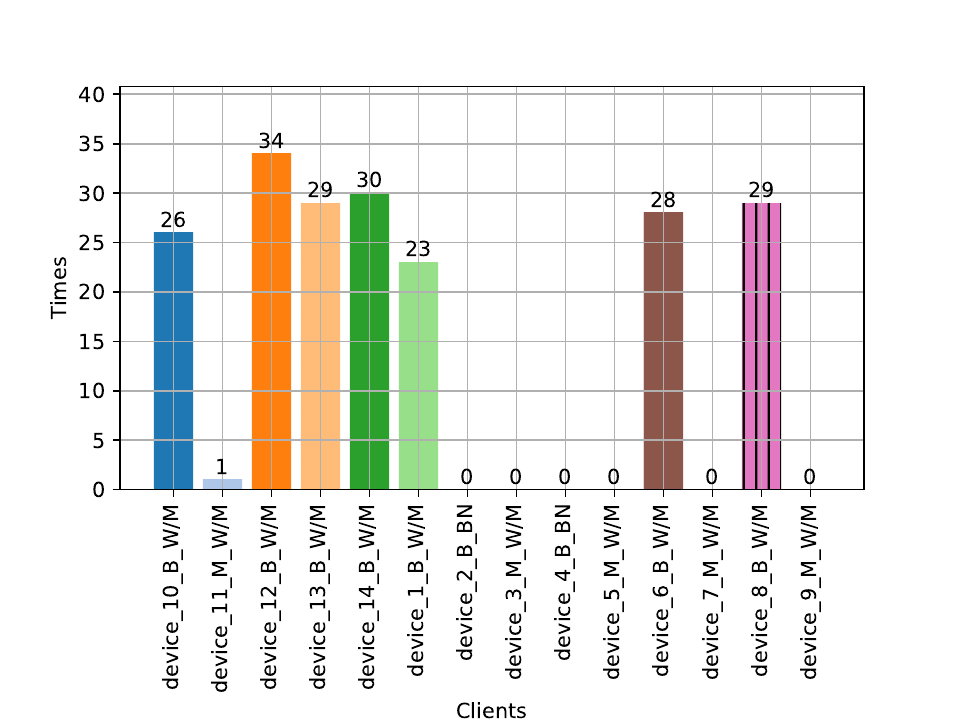}
	\caption{Fashion-MNIST-IID-Signflip}
	\label{fashionmnist-iid-signflip-50}
\end{subfigure}
\begin{subfigure}[b]{0.22\textwidth}
	\centering
	\includegraphics[width=\textwidth]{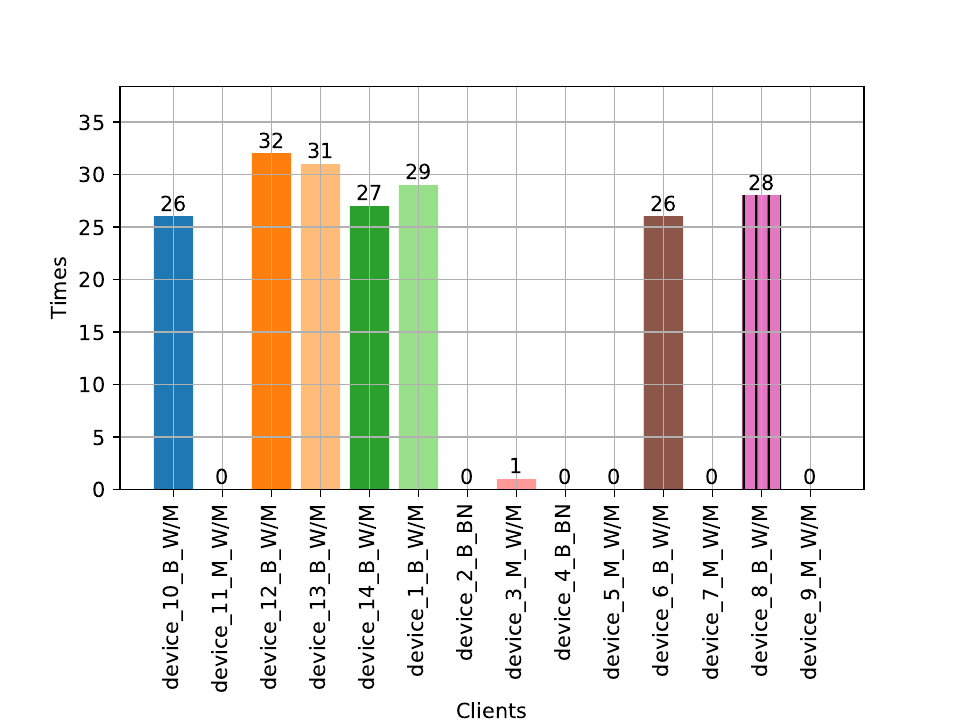}
	\caption{Fashion-MNIST-Non-IID-Noise}
	\label{fashionmnist-non-iid-noise-50}
\end{subfigure}
\begin{subfigure}[b]{0.22\textwidth}
	\centering
	\includegraphics[width=\textwidth]{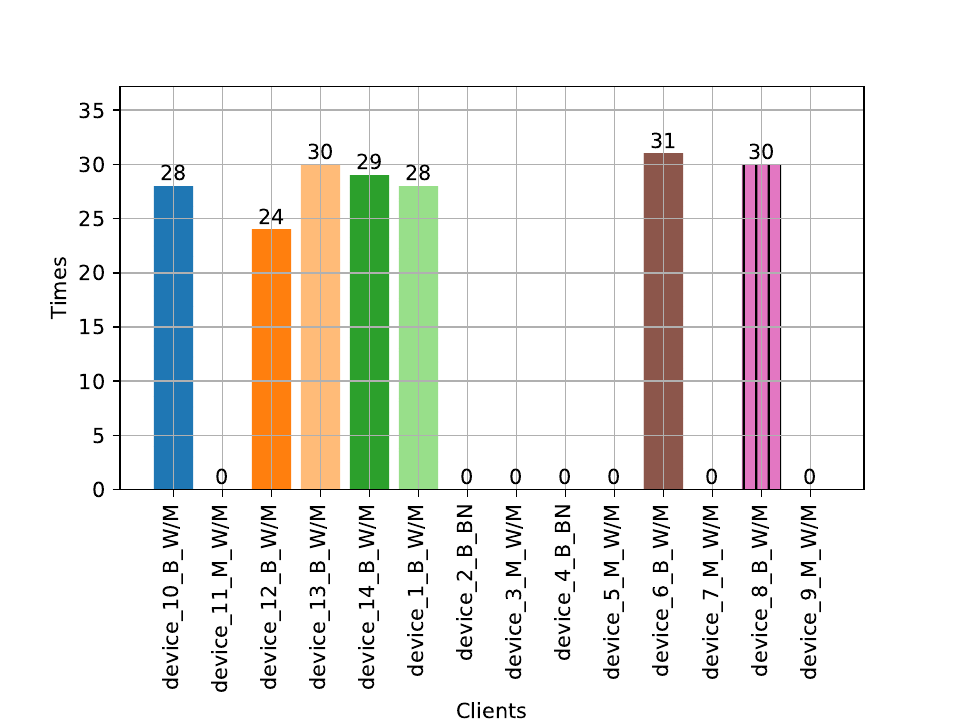}
	\caption{Fashion-MNIST-Non-IID-Signflip}
	\label{fashionmnist-non-iid-signflip-50}
\end{subfigure}

\caption{Frequency of nodes in  BRLF as aggregation nodes under 50\% attack rate with two types of byzantine attacks for MNIST and Fashion-MNIST tasks }
\label{being_miner_times}
\end{figure*}

In this section, we discuss the number of times local training nodes are selected as aggregation nodes using the PPCC consensus algorithm, as shown in Figure \ref{being_miner_times}, within the same environment as Figure \ref{Accuracy}, \ref{Stake}, and \ref{blockchain_size}.

Figure \ref{being_miner_times} is a bar chart that represents the number of times different clients are selected to become aggregation nodes during federated learning. The x-axis represents the different clients, and the y-axis represents the number of times a client has been selected as an aggregation node, with corresponding numbers labeled in the figure. 

As seen in Figure \ref{being_miner_times}, blockchain nodes, such as \textit{device\_2\_B\_BN} and \textit{device\_4\_B\_BN}, do not participate in the federated learning process, so their count is 0. Among the nodes of type \textit{W/M}, the number of times benign nodes become aggregation nodes is relatively evenly distributed, without excessive or under-representation. This is due to the fact that the PPCC algorithm does not allow consecutive selection as an aggregation node each time.

In most cases, the number of times malicious nodes become aggregation nodes is 0. However, there are still a small number of cases where they become aggregation nodes. This may occur when there is no global model at the initialization of federated learning or when the global model is not yet perfect, resulting in random selection of nodes as aggregation nodes during the first round. However, as observed from the figure, the occurrences of malicious nodes becoming aggregation nodes are minimal. This indicates that the PPCC algorithm effectively prevents malicious nodes from being selected as aggregation nodes.

\section{Conclusion}
\label{Conclusion}


In this work, we propose BRFL, a blockchain-based Byzantine-robust federated learning model. Our objective is to address the issues of \textit{byzantine-attack} and \textit{resource-cost} during model aggregation. The \textit{resource-cost} problem can be further divided into computation resource costs and storage resource costs. To achieve this, we introduce two key components: the Proof of Pearson Correlation Coefficient (PPCC) consensus algorithm and the Precision-based Spectral Aggregation (PSA) algorithm. These components verify the accuracy of the average gradient by clustering the local model gradients and store federated learning communication records in the blockchain. Additionally, we utilize IPFS to store the local models and allow nodes to selectively store blocks, effectively managing storage requirements. Experimental results demonstrate that BRFL maintains Byzantine robustness with reduced resource costs.

\section*{Acknowledgments}
This should be a simple paragraph before the References to thank those individuals and institutions who have supported your work on this article.




\section{Biography Section}
 
\vspace{11pt}

\vspace{-33pt}
\begin{IEEEbiography}[{\includegraphics[width=1in,height=1.25in,clip,keepaspectratio]{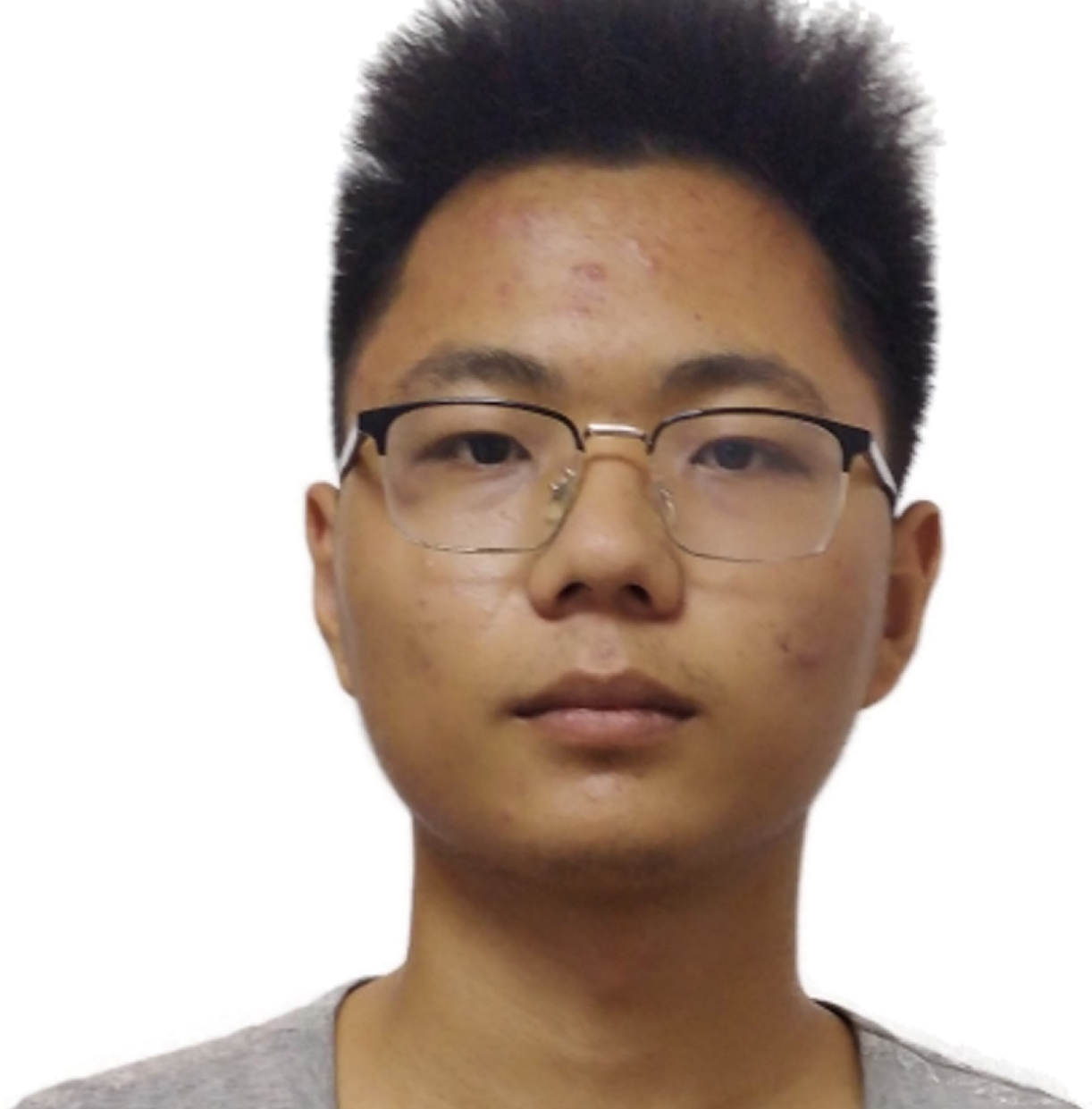}}]{Yang Li}
 received the master’s degree in computer technology from  Zhengzhou University, Zhengzhou, China, in 2021. He is
currently pursuing the Ph.D. degree with Beihang
University, Beijing, China, under the supervision of
Prof. Chunhe Xia.
His research interests include  federated learning and blockchain.
\end{IEEEbiography}


\begin{IEEEbiography}[{\includegraphics[width=1in,height=1.25in,clip,keepaspectratio]{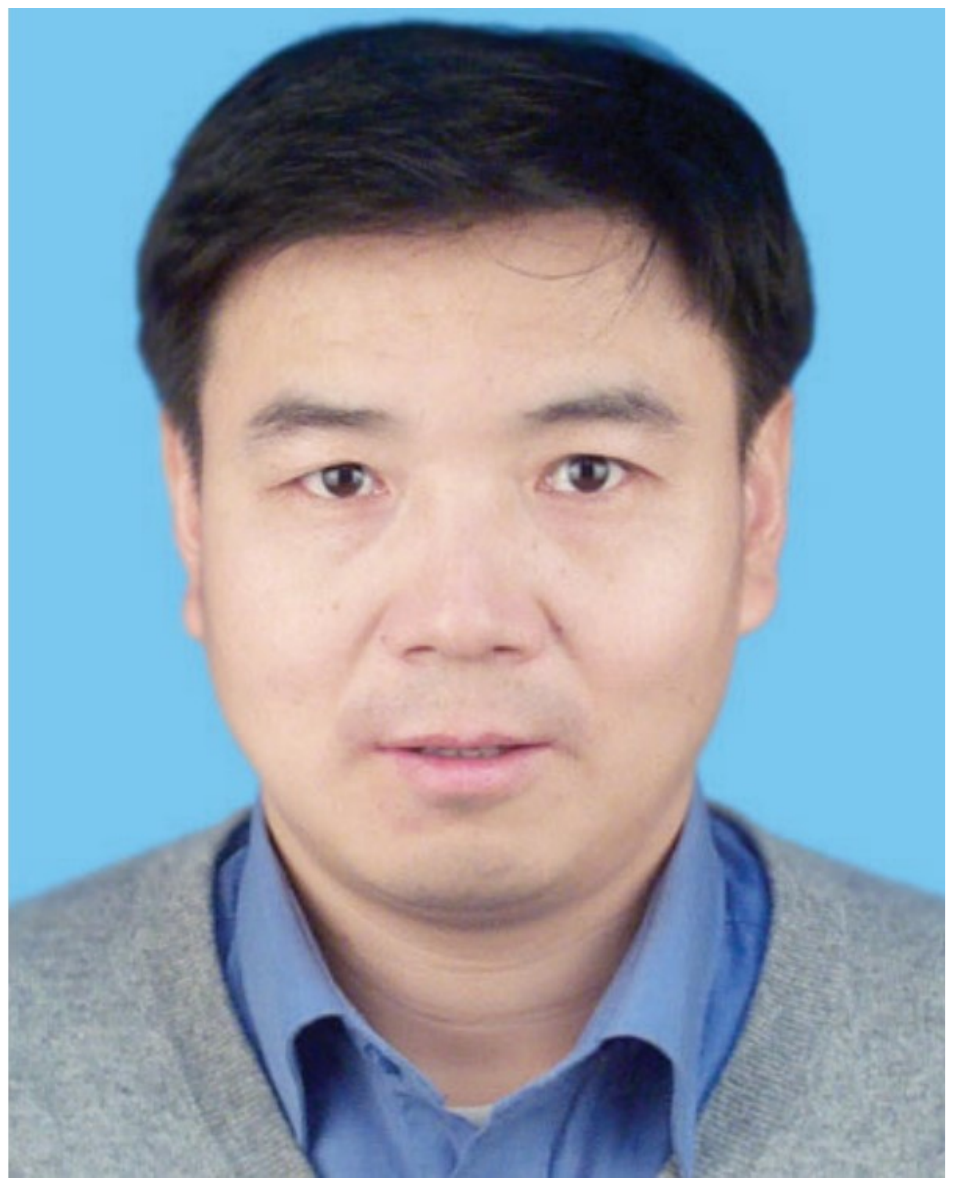}}]{Chunhe Xia}
	received the Ph.D. degree in computer application from Beihang University, Beijing, China, in 2003.
	
	He is currently a Supervisor and a Professor with Beihang University, where he is also the Director of the Beijing Key Laboratory of Network Technology and a Professor of the Guangxi Collaborative Innovation Center of Multi-Source Information Integration and Intelligent Processing. He has participated in different national major research projects and has published more than 70 research papers in important international conferences and journals. His current research focuses on network and information security, information 	countermeasure, cloud security, and network measurement.
\end{IEEEbiography}

\begin{IEEEbiography}[{\includegraphics[width=1in,height=1.25in,clip,keepaspectratio]{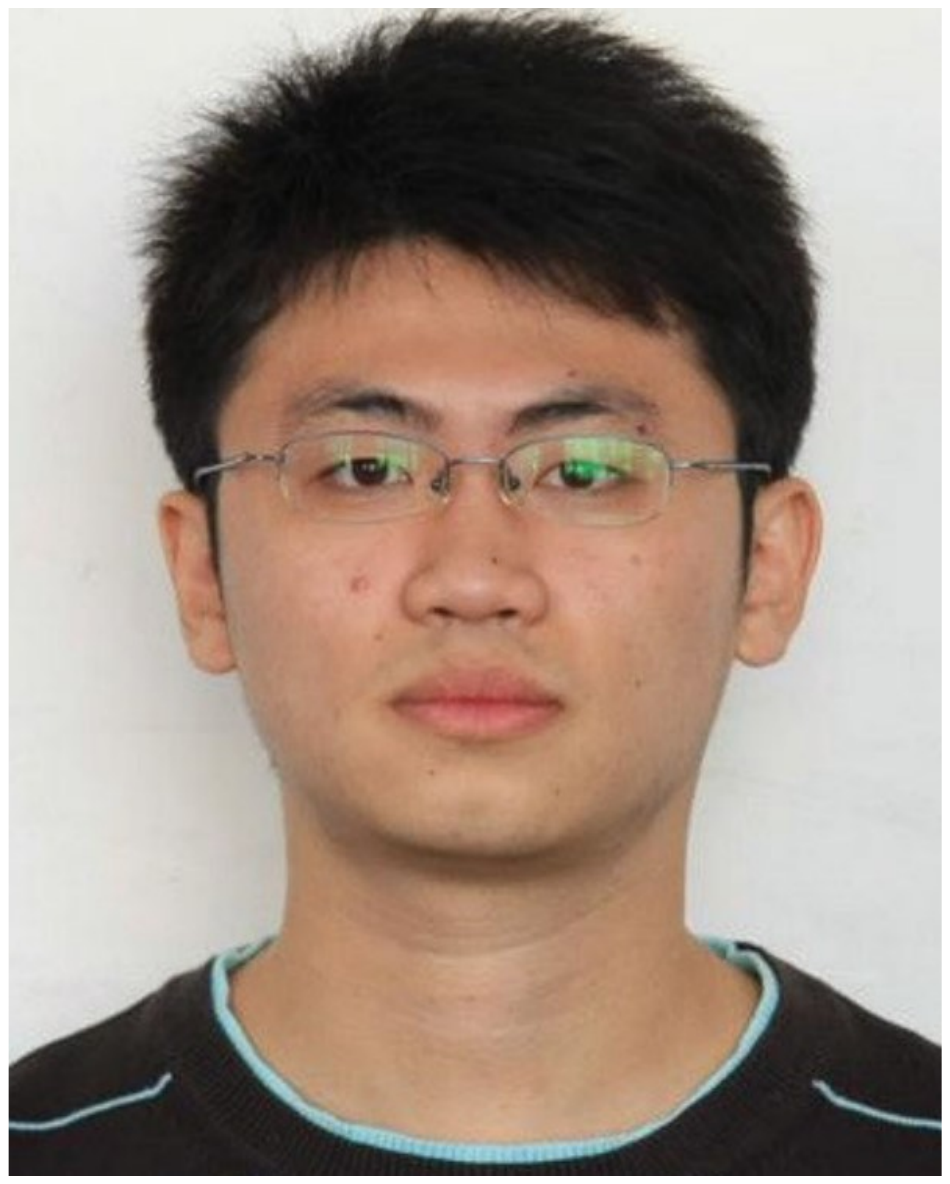}}]{Tianbo Wang}
(Member, IEEE) received the Ph.D. degree in computer application from Beihang University, Beijing, China, in 2018.

He is currently an Associate Professor with Beihang University, where he is also an Associated Professor of the Shanghai Key Laboratory of Computer Software Evaluating and Testing. He has participated in several National Natural Science Foundations and other research projects. His research interests include network and information security, intrusion detection technology, and information countermeasure.
\end{IEEEbiography}

%

\vfill

\end{document}